\documentclass[twoside,11pt]{article}

\usepackage{jmlr2e}

\usepackage{booktabs}
\usepackage{listings}
\usepackage{amsmath}
\usepackage{comment}
\usepackage{lineno,hyperref}
\usepackage{amssymb}
\usepackage{amsmath}
\usepackage{verbatim}
\usepackage[ruled,vlined]{algorithm2e}

\usepackage{float}
\usepackage{xfrac}
\usepackage[all]{xy}
\usepackage{siunitx}
\usepackage{color}
\usepackage{nth}
\usepackage{graphicx}
\usepackage{pgfplots}
\pgfplotsset{compat=1.14}

\pgfplotsset{compat=newest}
\pgfplotsset{plot coordinates/math parser=false}
\usepackage{tikzscale}
\usetikzlibrary{matrix,chains,positioning,decorations.pathreplacing,arrows}
\usepackage{tikz-qtree,tikz-qtree-compat}
\usetikzlibrary{calc}

\newcommand{\ts}{\textsuperscript}

\ShortHeadings{Explainable Artificial Intelligence in Digital Pathology}{Holzinger, Malle, Kieseberg, Roth, M\"uller, Reihs, Zatloukal}
\firstpageno{1}

\begin{document}

\title{Towards the Augmented Pathologist:\\
Challenges of Explainable-AI in Digital Pathology}


\author{\name Andreas Holzinger\textsuperscript{1} \email a.holzinger@hci-kdd.org\\
\name Bernd Malle\textsuperscript{1,2}  \email b.malle@hci-kdd.org\\
\name Peter Kieseberg\textsuperscript{1,2,3} \email p.kieseberg@sba-research.org\\
\name Peter M. Roth\textsuperscript{4}\email pmroth@icg.tugraz.at\\
\name Heimo M\"uller\textsuperscript{1,5}\email heimo.mueller@medunigraz.at\\
\name Robert Reihs\textsuperscript{1,5}\email robert.reihs@medunigraz.at\\
\name Kurt Zatloukal\textsuperscript{5}\email kurt.zatloukal@medunigraz.at\\
\\
\addr \textsuperscript{1} Institute for Medical Informatics, Statistics\,\&\,Documentation,\,Medical\,University\,Graz,\,Austria\\
\addr \textsuperscript{2} Secure Business Austria, SBA Research gGmbH, Vienna, Austria\\
\addr \textsuperscript{3} University of Applied Science St. P\"olten, Austria\\
\addr \textsuperscript{4} Institute of Computer Graphics and Vision, Graz University of Technology, Austria\\
\addr \textsuperscript{5} Institute of Pathology, Medical University Graz, Austria\\
}


\maketitle

\begin{abstract}
Digital pathology is not only one of the most promising fields of diagnostic medicine, but at the same time a hot topic for fundamental research. Digital pathology is not just the transfer of histopathological slides into digital representations. The combination of different data sources (images, patient records, and *omics data) together with current advances in artificial intelligence/machine learning enable to make novel information accessible and quantifiable to a human expert, which is not yet available and not exploited in current medical settings. The grand goal is to reach a level of usable intelligence to understand the data in the context of an application task, thereby making machine decisions transparent, interpretable and explainable. The foundation of such an "augmented pathologist" needs an integrated approach: While machine learning algorithms require many thousands of training examples, a human expert is often confronted with only a few data points. Interestingly, humans can learn from such few examples and are able to instantly interpret complex patterns. Consequently, the grand goal is to combine the possibilities of artificial intelligence with human intelligence and to find a well-suited balance between them to enable what neither of them could do on their own. This can raise the quality of education, diagnosis, prognosis and prediction of cancer and other diseases. In this paper we describe some (incomplete) research issues which we believe should be addressed in an integrated and concerted effort for paving the way towards the \textit{augmented pathologist.}
\end{abstract}


\section{Introduction}
\label{sec01}

Artificial intelligence (AI) is currently experiencing a tremendous hype towards building systems that can learn and think like humans do \citep{LakeUlmanTenenbaumGershman2016:MachinesThinkArXiv,WeberEtAl:2017:AugmentedAgentsDeep}. Much of the current success results from the applicability of deep neural networks trained on large data sets in various tasks. These include object recognition, e.g. identifying cats from unlabeled images \citep{LeCorradoDeanNg:2011:theCat}, video games, e.g. playing Atari games \citep{MnihSilverEtAl:2013:AtariDeepLearningArXiV}, or board games, e.g. the game of Go \citep{HassabisEtAl:2016:GoNature}. These systems achieve a performance that equals or even beats humans in some respects~\citep{SilverHassabisEtAl:2017:GoWithoutHuman}.

An impressive example from the medical domain is the very recent work by the Thrun-Group \citep{EstevaThrun:2017:DermaNN}, which showed that convolutional neural networks can achieve a performance on par with human doctors, demonstrating the ability of classifying skin cancer with a level of competence comparable to dermatologists. Despite all these impressive successes of automatic approaches we are still very far away from reaching general \emph{human-level AI}. This would require algorithms to deal with the common sense "informatic" situation - we say context -  in which the phenomena to be taken into account in achieving goals are not yet fixed in advance \citep{McCarthy:2007:HumanLevelAI}.

In our opinion, there is no danger that AI will replace medical professionals in the medium term; however, there is an increasing trend that those who will not enrich their methods with AI will have a potential disadvantage to those who embrace cutting-edge Machine Learning (ML) technologies.

AI and ML are often used synonymously, although there is a difference: AI is the field working on understanding intelligence and encompasses all underlying scientific theories of human intelligence/human learning versus machine intelligence/machine learning. ML is a very practical field and applies the findings of AI research to the design, development and testing of algorithms that can learn from data, gain knowledge from experience and improve their learning behavior over time \citep{Holzinger:2017:InauguralMAKE}. Understanding human intelligence, i.e. to study how humans make decisions and how they reason, is difficult but of utmost importance for advances in AI and ML. For a number of reasons it is very difficult to deal with non-stationary, non-linear as well as non-independent \& non-identically distributed data \footnote{Learning from non-iid data is difficult, \citep{ZhangEtAl:2009:nonIID}} in high-dimensional spaces, often additionally dependent on time. This is quite often the case in the medical domain where we are confronted with a variety of different data (for a taxonomy of data refer to \citep{HolzingerDehmer:2014:TaxonomyData,Holzinger:2014:SpringerTextbook}).

Studying the work of pathologists is interesting for several reasons: 1) Digital pathology is not just the transformation of the classical microscopic analysis of histological slides by pathologists to a digital visualization, it is an innovation that will dramatically change medical workflows in the coming years; 2) Much information is hidden in arbitrarily high dimensional spaces, not accessible to a human, consequently we need an AI/ML approach to generate a new kind of information, which is not yet available and not exploited in current diagnostics; 3) Pathologists are able to transfer previously learned knowledge quickly to new tasks. Insights into the latter supports AI research generally and ML research specifically and may contribute to developing software which can learn from experience, similarly as we humans do. Technologically, during the last decade pathology has benefited from the rapid progress of image digitizing technologies, which led to the development of scanners capable to produce Whole Slide images (WSI) which can be explored by a pathologist on a computer screen (virtual microscope) comparable to the conventional microscope and can be used for education and training, diagnostics (clinico-pathological meetings, consultations, revisions, slide panels and upfront clinical diagnostics) and archiving \citep{AlJanabiEtAl:2012:DigitalHistopathology}. Practical implications of digital pathology include telepathology \citep{AfeworkEtAl:1998:Telepathology}, second opinions \citep{EpsteinEtAL:1996:SecondOpinion} and education \citep{Dee:2009:VirtualMicroscopyEducation}.

The digitalization opens new opportunities to extract knowledge from such data, which would not be accessible to the human expert otherwise \citep{HolzingerEtAl:2014:KDDBio}; however, standard image analysis tools are inadequate to deal with the specialities of WSI's.

Fusing traditional image sources with *omics data calls for new solutions \citep{MadabhushiLee:2016:MLDigitalPathology}, particularly if we want to go towards personalized medicine \citep{Holzinger:2014:trends}.
The variety of problems in digital pathology requires a synergistic combination of various methodological approaches which calls for an integrated ML approach \citep{Holzinger:2012:DATAconf,Holzinger:2013:HCI-KDD,HolzingerEtAl:2017:DigitalPathologyMachineLearning}. Methodologies of two areas offer ideal conditions to implement integrated ML: Human--Computer Interaction (HCI) and Knowledge Discovery/Data Mining (KDD), with the goal of augmenting human intelligence with artificial intelligence to discover novel, previously unknown insights into data. This calls for solutions which let the medical professional not only ask questions like "Where are similarities/differences/anomalies ...?" but to ask \textit{"why" questions,} for example "Why are there similarities/differences/anomalies ...?" in order to find and explain unknown unknowns in complex data, which goes beyond data mining.
Interestingly, early developments from the field of augmenting human intelligence came from HCI-research \citep{EngelbartEnglish:1968:StanfordCenter,Engelbart:1995:Augmenting,RusselNorvig:1995:AIbook}.

\section{From Microscopical Thinking to Digital Pathology}
\label{sec02} \label{sec:digitalpathology}

\begin{figure}[ht!]
\centering
\includegraphics[width=\textwidth]{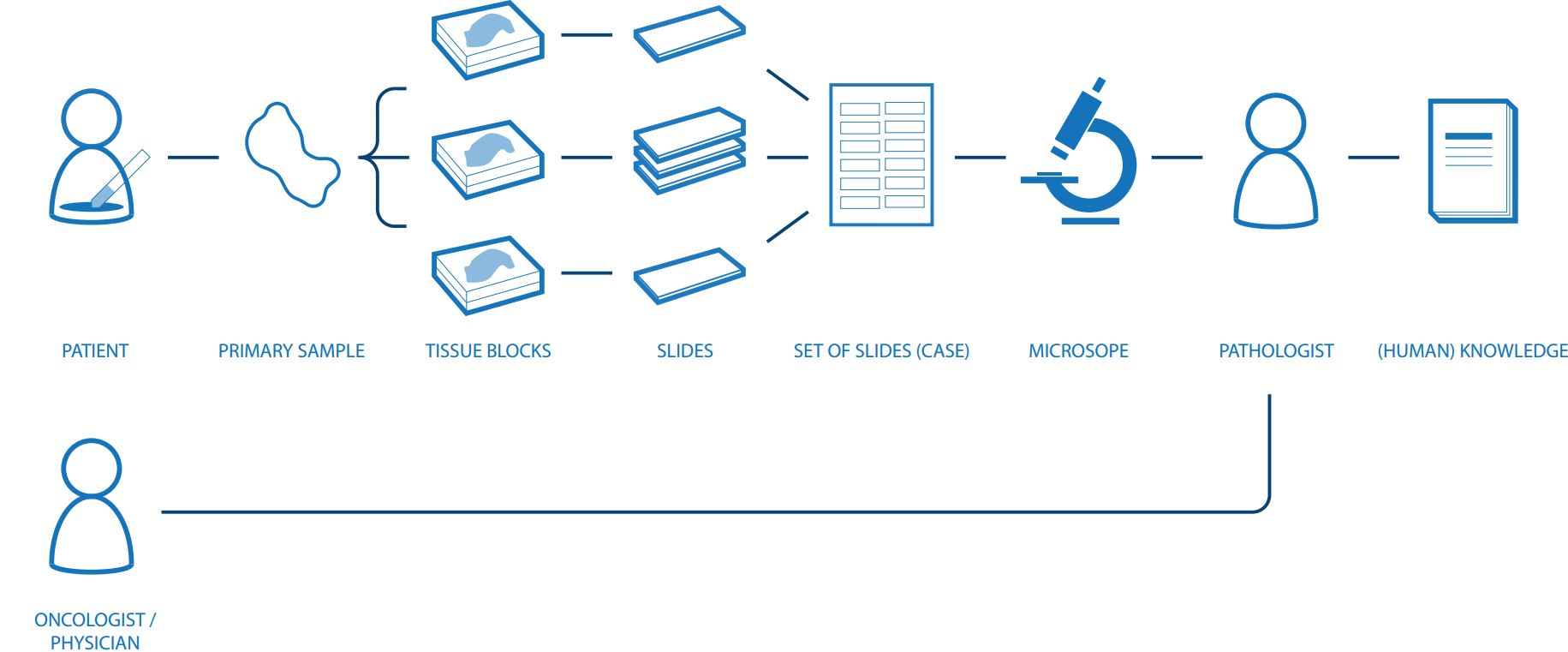}
\caption{Clinical pathology, from mid $19^{\text{th}}$ century up to the end of the $20^{\text{th}}$ century.}
\label{fig:classicalpathology}
\end{figure}

Modern pathology was founded in the mid of the $19^{\text{th}}$ century, when \cite{Virchow:1871:Patho} set the basis for modern medical science and established the "microscopical thinking" which is still state-of-the-art in classical pathology. In histopathology a biopsy or surgical specimen is examined by a pathologist, after the specimen has been processed and histological sections have been placed onto glass slides, see Figure \ref{fig:classicalpathology}. In cytopathology either free cells (fluids) or tissue micro-fragments are "smeared" on a slide without cutting any tissue. An alternative to invasive autopsy is virtual autopsy which is also conducted with scanning and imaging techniques \citep{Shane:2017:VirtualAutopsyFirst}.

\begin{figure}[ht!]
\centering
\includegraphics[width=\textwidth]{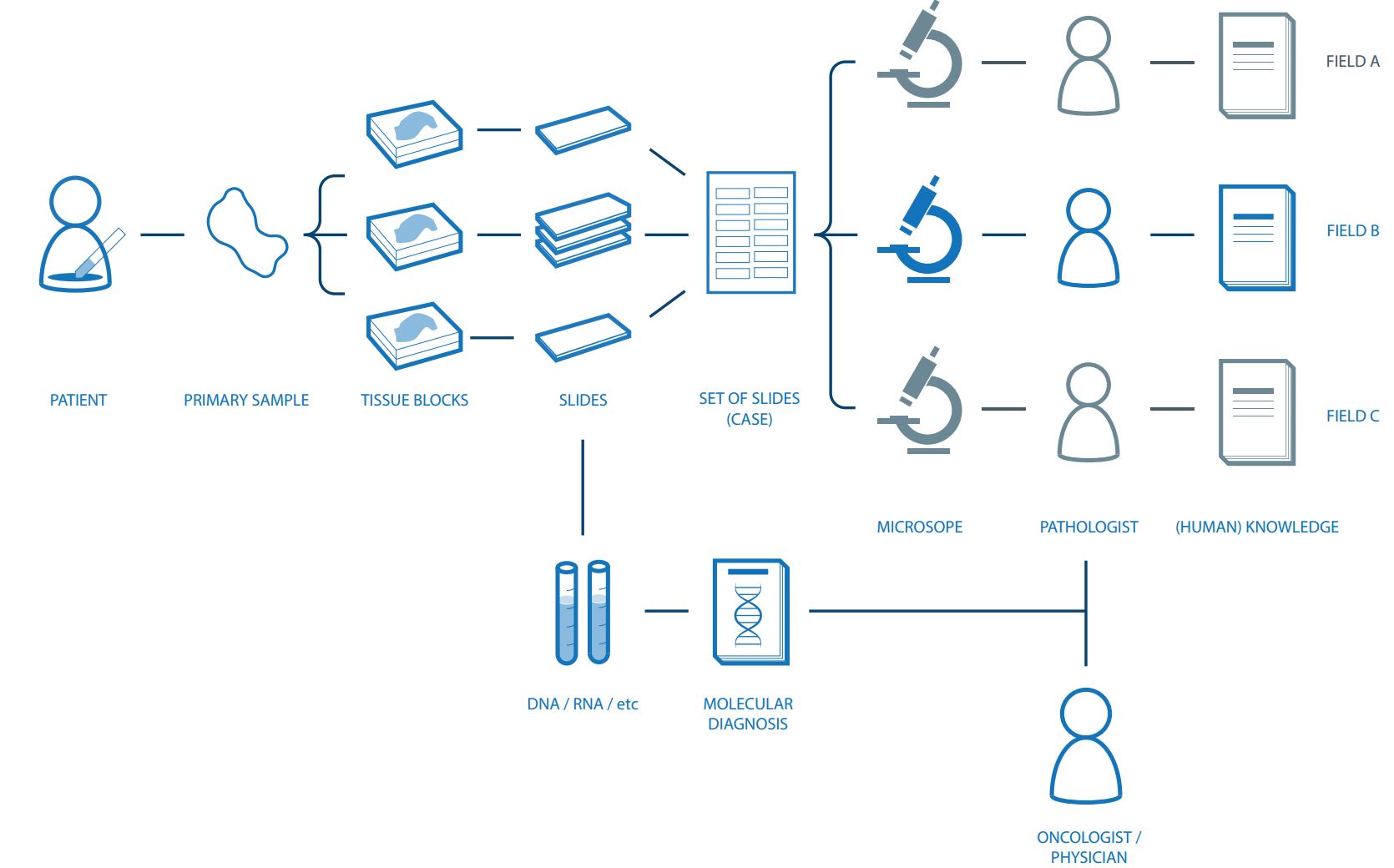}
\caption{Today's state-of-the-art (non digital) clinical pathology.}
\label{fig:modernpathology}
\end{figure}

Towards the end of the $20^{\text{th}}$century an individual clinical pathologist was no longer able to cover the knowledge of the whole scientific field. This led to the specialization of clinical pathology either by organ systems or methodologies. Molecular biology and *omics technologies set the foundation for the emerging field of molecular pathology, which today alongside WSI provides the most important source of information, especially in the diagnosis of cancer and infectious diseases, see Figure \ref{fig:modernpathology}.

The roots of digital pathology go back to the 1960s in the context of telepathology, which was a remote-controlled light microscope attached to a video camera and a telecommunication link \citep{Weinstein:1987:telepathology}.

\begin{figure}[ht!]
\centering
	\includegraphics[width=\textwidth]{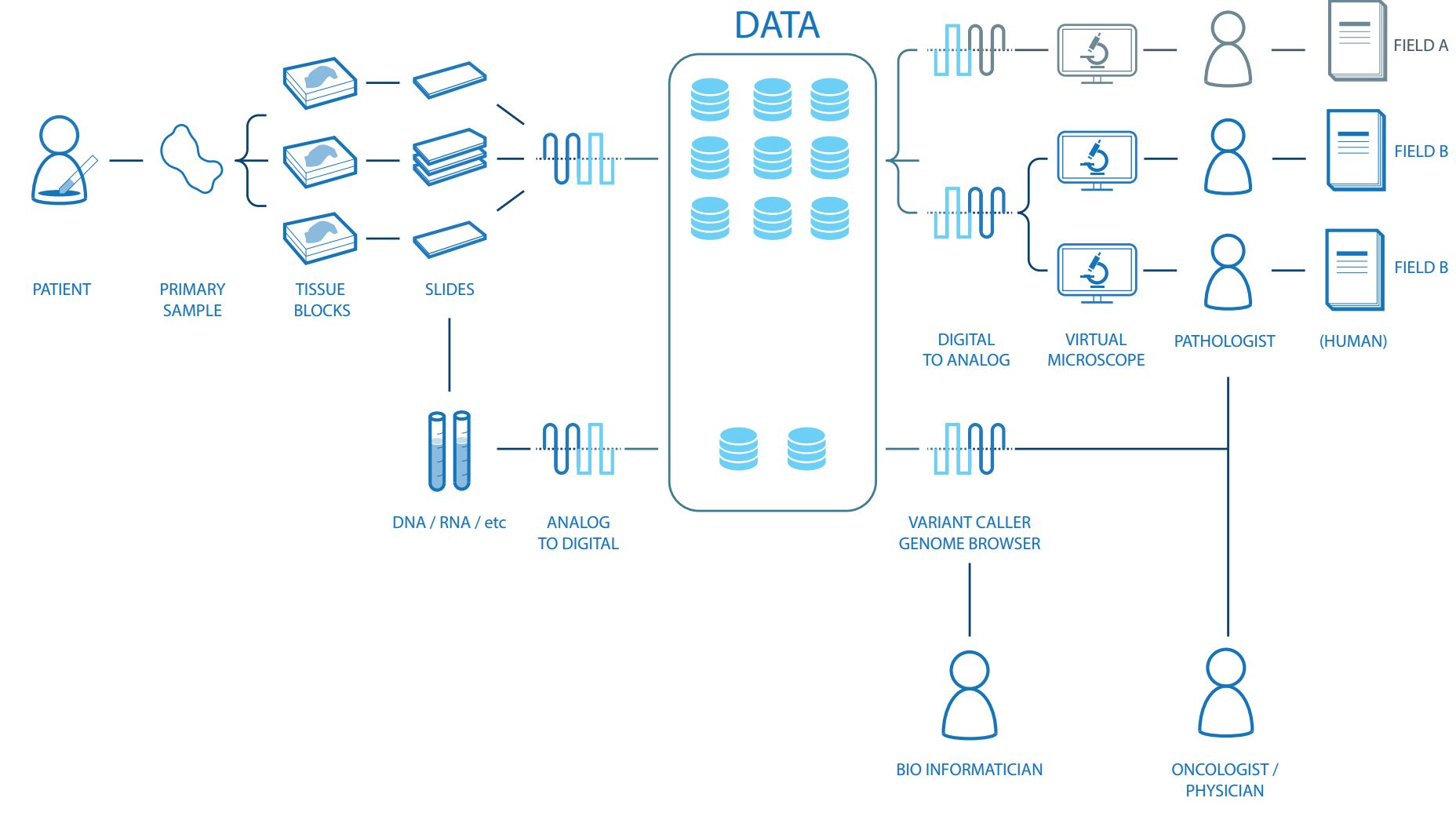}
\caption{A digital pathology workflow always starts with the gross evaluation of the primary sample. Depending on the medical question and the material type, small tissue parts are extracted from the primary sample. The pathology labs cuts several slides from the tissue blocks, applies different staining methods and conducts additional histological and molecular tests. In addition to the set of whole slide images (each about 16 Gigapixels in size!), textual information from medical reports as well as molecular data are combined in the diagnosis-making process.}
\label{fig:digitalpathology}
\end{figure}

Later in the 1990s the principle of virtual microscopy \citep{WeinsteinEtAl:2009:VirtualMicroscopy} appeared in several life science research areas. At the turn of the century the scientific community gradually adopted the term  "digital pathology" \citep{BarbareschiEtAl:2000:DigitalPathology} to  denote digitization efforts in pathology, see Figure \ref{fig:digitalpathology}.

The technical requirements (scanner, storage, network) in digital pathology are magnitudes higher than in radiology due to some general differences between those two fields: In radiology the image is primarily captured in digital format, whilst in pathology the scanning is done from preserved and processed specimens, for retrospective studies even from slides stored in biobanks \citep{Huppertz:2014:Biobank}. The integration of biobanks is of enormous importance for the realization of future precision medicine \citep{MuellerHolzinger:2015:BiobankIntegration}.

Besides some differences in pre-analytics workflows and metadata content, the required storage in digital pathology is two to three orders of magnitude higher than in radiology and places heavy demands on IT infrastructure \citep{HuismanEtAl:2010:DigitalPathology}.

\subsection{Virtual Case}
\label{ssect:virtualcase}

A pathological workflow always starts with the overall evaluation of the primary sample. Depending on the medical question and the material type, small tissue parts are extracted from the primary sample and are either embedded in a paraffin block or cryo-frozen. From the tissue blocks the pathology lab cuts several slides, applies different staining methods and conducts additional histological and molecular tests.  Finally, the pathologists evaluate all the slides together with the supporting gross- and molecular findings and makes the diagnosis. If in addition to the set of WSI all information is present in a structured digital format, we call this a virtual case. In a virtual case, the average number of slides and additional findings varies very much for different medical questions and material types.

The most demanding data elements in a virtual case are the whole slide images (WSI). Compared to radiology, where the typical file sizes are in the range from 500 KB to 50 MB, a single WSI scan with 80x magnification consists of approximately 16 Gigapixels (Note: for the calculation of the WSI file size and comparison of different scanner manufacturers, we use the de-facto standard area of 15mm x 15mm, with an optical resolution of \SI{0.12}{\micro\metre}, which corresponds to an 80x magnification).

With 8bit information for each color channel one WSI results in approx. 50GB stored in an uncompressed image format. Looking at the number of slides of a typical case, it is clear that some compression techniques must be applied to the image data, and luckily several studies reported that lossy compression with a high quality level does not influence the diagnostic results; Still there are unresolved questions \citep{HolzingerEtAl:2017:DigitalPathologyMachineLearning}.

The newest generation of scanners (status as of October 2017\footnote{As of October 2017 the worldwide fastest WSI scanner is working in Graz https://goo.gl/yiVnEx}) is able to digitize a slide at various vertical focal planes, called z-layers, each the size of a singe layer. The multi-layer image can be either combined by algorithms to a single composite multi-focus image (Z-stacking) or used to simulate the fine focus control of a conventional microscope. Z-stacking is a desirable feature especially when viewing cytology slides, however the pathologist should be aware that such an image can never be seen through an actual microscope.

At the Institute of Pathology of the Medical University of Graz about 73,000 diagnoses are made per year and approximately 335,000 glass slides are produced in the pathology lab; additionally approximately 25,000 glass slides are generated in the cytology lab. This results in a required yearly storage capacity of almost 1 PB and the appropriate computing power to process approximately 1000 slides per day plus the necessary capacity to train and improve AI/ML algorithms.

Several data formats are used today, either vendor independent  (DICOM, TIFF/BigTIFF, Deep Zoom images) and vendor specific formats from Aperio, Hamamatsu, Leica, 3DHistech, Philips, Sakura and Trestle. In the setup of a virtual slide archive for medical research and machine learning it is essential to a) agree on a common exchange format, and b) to separate patient related and image related metadata. Patient related metadata comprise direct identifiers (name, birthday, zip code, ...) but also diagnosis results and others results from the patient medical history. When no such data is stored within or attached to the image format, the WSI is purely anonymous as no re-identification of the patient is possible. To link between the same WSI used in different studies, either a global unique identifier (GUID) or an image generated hash can be used.

\subsection{Towards the Augmented Pathologist}
\label{subsec:augmentedpathologist}

Digitizing the workflows is one important fundamental issue to enable a revolutionary change in clinical pathology. Based on the digital workflow AI paradigms and ML methods can be applied to augment the human pathologist during diagnosing, research, education and training. The augmentation may start with simple classification and quantification algorithms as already available today (e.g. cell sorting \citep{ArtetaZisserman:2012:Celldetection}), and may end in a fully autonomous pathologist. This is similar to the concept of autonomous driving, where human intervention is no longer necessary, but can still range over various degrees of involvement (to-be-or-not-to-be-in-the-loop \citep{LouwEtAl:2015:DriverLoop}).

In driving automation autonomy levels have already been established (SAE J3016 \footnote{Taxonomy and Definitions for Terms Related to On-Road Motor Vehicle Automated Driving Systems. Society of Automotive Engineers (SAE, 2014).}) \cite{KircherEtAl:2014:levelsAutomation}: level 0 = full control of the driver, just warnings; level 1 = hands-on, driver must be ready to retake full control at any moment; level 2 = hands-off, driver must monitor at any time and be prepared to immediately intervene in case the automatic system fails; level 3 = eyes-off, driver can safely turn her/his attention elsewhere, but driver may take over at any time; level 4 = mind-off, driver even may go to sleep or leave the driver's seat; level 5 = fully automatic, no human intervention required, e.g. a robotic taxi.

To distinguish such extreme scenarios from simple digital workflows we propose the term {\bfseries machine aided pathology}, meaning that significant contributions of the decision making process are supported by machine intelligence. Such machine aided pathology solutions can be applied at several steps during the diagnosis and decision making process:

\begin{description}
\item[Formulation of a hypothesis.] {
  Each diagnosis starts with a medical question and a corresponding underlying initial hypothesis. The pathologist refines this hypothesis in an iterative process, consequently looking for known patterns in a systematic way in order to confirm, extend or reject his/her initial hypothesis.
Unconsciously, the pathologist asks the question \textit{"What is relevant?"} and zooms purposefully into the - according to his/her opinion - essential areas of the cuts. The duration and the error rate in this step vary greatly between inexperienced and experienced pathologists. An algorithmic support in this first step would contribute in particular to the quality and interoperability of pathological diagnoses and reduce errors at this stage, and would be particularly helpful for educational purposes. A useful approach is known from \cite{ReederFelson:2003:Gamuts} to discover certain patterns (called gamuts) in radiological images and to classify these into the groups ''common/uncommon''. This approach has its roots in differential diagnosis which has a long tradition \citep{Kobayashi:1974:differentialDiagnosis} and has been implemented with artificial neural networks \citep{AsadaDoi:1997:PatentDD}.

- Very large amounts of data can only be managed with a "multi resolution" approach using image pyramids. For example, a Colon cancer case consists of approximately 20 Tera (!) pixels of data - a size which no human is capable of processing.

- The result of this complex process is a central hypothesis, which has to be tested on a selection of relevant areas in the WSI, which is determined by quantifiable values (receptor status, growth rate, etc.).

- Training data sets for ML can now contain human learning strategies (transfer learning, e.g.) as well as quantitative results (hypotheses, areas, questions, etc.).
}

\item[Detection and classification of known features.]{
Through a precise classification and quantification of selected areas in the sections, the central hypothesis is either clearly confirmed or rejected. In this case, the pathologist has to consider that the entire information of the sections is no longer taken into account, but only areas relevant to the decision are involved. It is also quite possible that one goes back to the initial hypothesis step by step and changes their strategy or consults another expert, if no statement can be made on the basis of the classifications.

- In this step ML algorithms consist of well known standard classification and quantification approaches.

- An open question is how to automatically or at least semi-automatically produce training sets, because here specific annotations are needed (which could come from a stochastic ontology, e.g.).

- Another very interesting and important research question is, whether and to what extent solutions learned from one tissue type (organ 1) can be transferred to another tissue type (organ 2) -- transfer learning -- and how robust the algorithms are with respect to various pre-analytic methods, e.g. stainings, etc.
}

\item[Risk prediction and identification of unknown features.]
Within the third step, recognized features (learned parameters) are combined into a diagnosis and an overall prediction of survival risk. The main challenge in this step lies in training/validation and in the identification of novel, previously unknown features from step two. We hypothesize that the pathologist supported by machine learning approaches is able to discover patterns -- which previously were not accessible! This would lead to totally new insights into previously unseen or unrecognized relationships.

\end{description}

Beyond challenges in ML, the following general topics and prerequisites have to be solved for a successful introduction of machine aided pathology:

\begin{description}
\item[Standardization] of WSI image formats and harmonization of annotation / metadata formats. This is essential for telepathology applications (on the importance of standardization for digital pathology see the recent work of \cite{BarisoniEtAl:2017:Standardization}); and even more important for the generation of training sets, as for a specific organ and disease stages, the required amount of cases may not be available even at a large institute of pathology.

\item[Common digital cockpit] and visualization techniques should be used in education, training and across different institutes. Changing the workplace should be as easy as switching the microscope model or manufacturer. However, commonly agreed-upon visualization and interaction paradigms can only be achieved in a cross vendor approach and with the involvement of major professional associations.
\end{description}

\subsection{Data Integration}
\label{subsec:dataintegration}

Applying sophisticated AI/ML methods on images alone is one aspect \citep{LiuStumpe:2017:GooglePatho}; however, the real added value results from the combination of two other sources of data:\\

\textbf{1) Clinical data from electronic patient records (EPR)} including the patient history with all documents, diagnoses, medical reports, laboratory tests, physiological parameters, vital signs, but also recorded signals, ECG, EEG, etc.); this also links to other image data including standard X-ray, MR, CT, PET, SPECT, microscopy, confocal laser scans, ultrasound imaging, molecular imaging, etc.\footnote{Remark: At Graz University Hospital an all-digital hospital information system is in operation since 2000 \citep{GellEtAl:2000:HospitalInformationSystem,HolzingerEtAl:2000:IntelligentTutoring}}

\textbf{2)*omics data from biobanks} \citep{Huppertz:2014:Biobank,OmmenZatloukal:2015:BBMRI,HainautPasterk:2017:Biobanking}, e.g. from genomic sequencing technologies (Next Generation Sequencing, NGS, etc.), microarrays, transcriptomic technologies, proteomic and metabolomic technologies, etc., which all play important roles for biomarker discovery and drug design \citep{McDermottEtAl:2013:Biomarker,SwanEtAl:2013:MLProteomics,LibbrechtNoble:2015:MLgenetics}.

\par Fusing and integrating this data (images, text, *omics) allows to make novel information accessible and quantifiable for the human pathologist, which are not yet available and not exploited to date. \textit{To understand the whole one must study the whole} (quote by \textsc{Henrik Kacser} (1986), reference in: \cite{Kell:2004:Sensemaking}). Data integration is a hot topic in the life sciences because solutions can bridge the gap between clinical routine and biomedical research \citep{JeanquartierEtAl:2016:OpenData}. This is becoming important due to the distributed, heterogeneous and diverse data sources, including picture archiving and communication systems (PACS) and radiological information systems (RIS), hospital information systems (HIS), lab information systems (LIS), physiological/clinical data repositories, and all sorts of *omics data in labs, and biobanks. It is of highest importance to provide end-users with a \textit{unified view on these data,}. These unified views are particularly important in high-dimensions, e.g. for integrating heterogeneous descriptions of the same set of genes \citep{LafonEtAl:2006:dataFusionML}, and fused data is more informative \citep{Blanchet:2016:DataFusionMetabolomics}.

\begin{figure}[ht!]
\includegraphics[width=\textwidth]{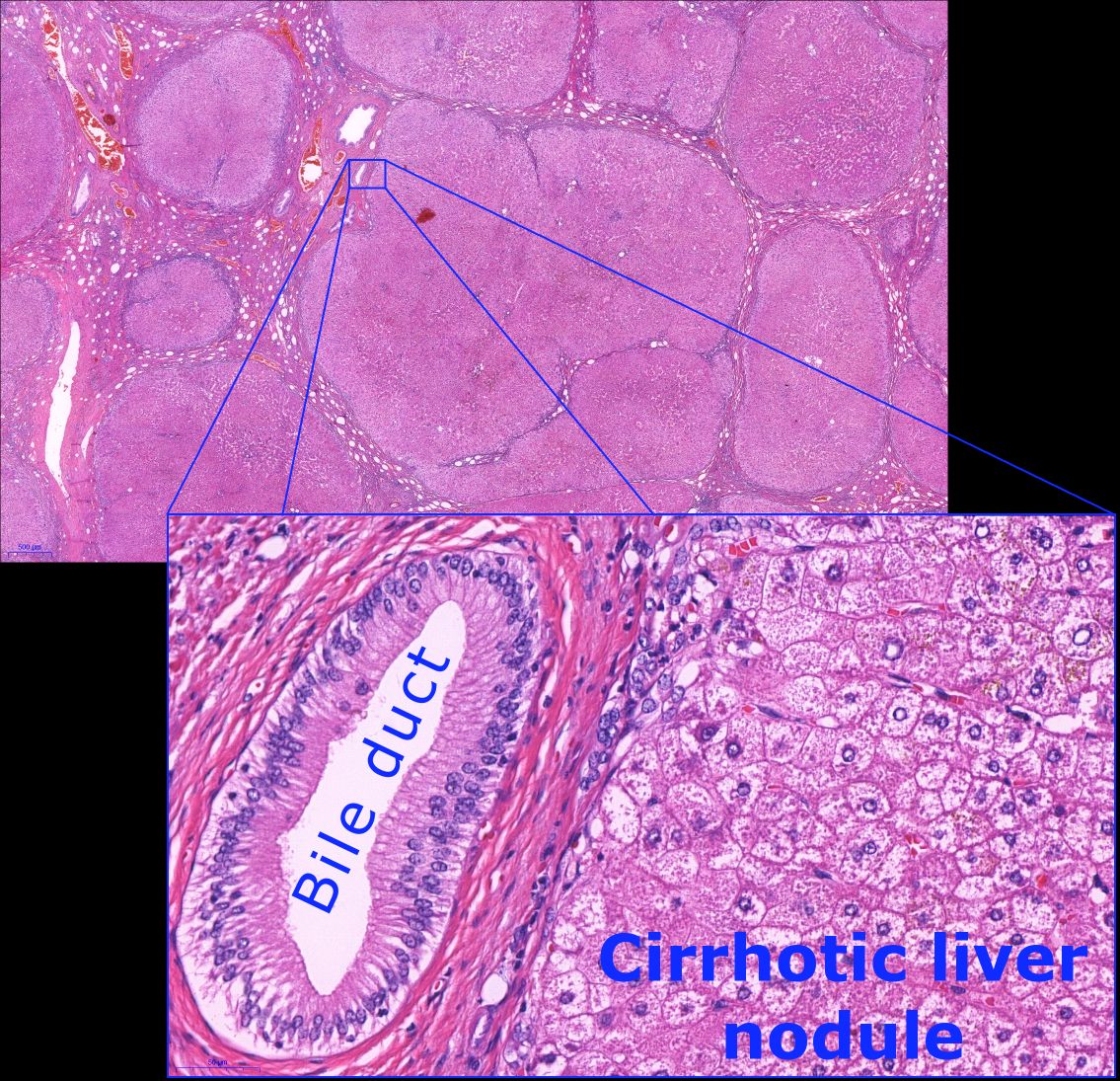}
\caption{Detail of a typical WSI: FFPE section of a human liver with cirrhosis, hematoxylin and eosin staining (Image Source: Pathology Graz, obtained with the new P1000 scanner of 3DHistech), see also Figure \ref{fig:carcinoma}}
\label{fig:cirrhosis}
\end{figure}


\newcommand{\argmin}{\operatornamewithlimits{arg\,min}}
\newcommand{\argmax}{\operatornamewithlimits{arg\,max}}
\newcommand{\trans}{^{\top}}
\newcommand{\st}{\text{s.t.~}}

\section{Deep Learning}
\label{sec:deeplearning}

Among the most important machine learning methods in medical image analysis are support vector machines, random forests, and deep learning approaches. For a brief overview of machine learning for digital pathology see \citep{HolzingerEtAl:2017:DigitalPathologyMachineLearning}. Here we briefly focus on the state-of-the-art in \textit{interpretable} deep learning. Deep Learning (DL, for an overview refer to \citep{LeCunBengioHinton:2015:DeepLearningNature,Schmidhuber:2015:DLOverview,ArelRoseKarnowski:2010:DLoverview}) is a whole family of (hierarchical) learning approaches primarily based on \textit{learning representations by layers and units in a network structure.} Due to their practical success DL is currently the most popular general framework \citep{Greenspan:2016:DeepMedicalImaging} in ML. The "deeper" the network the more layers and intra-layer units can represent functions of increasing complexity. Most tasks that consist of mapping an input vector to an output vector can be accomplished via deep learning, given sufficiently large data sets of labeled training examples (for details see the excellent book by \cite{GoodfellowBengioCourville:2016:DeepLearningBook}). Despite all their successes DL approaches have some limitations, e.g. they are very data hungry, need much computational power and they are considered black-box models \citep{SinghRibeiroGuestrin:2016:BlackBoxArxiv}. However, in the medical domain it is necessary to be able to open the black-box to a glass-box \citep{HolzingerEtAl:2017:glassbox} and to make the results transparent, re-traceable and explainable on demand (see section \ref{explainability}).

\subsection{Convolutional Neural Networks}
\label{subsec:cnn}

The most prominent and recently most successful DL architectures are Convolutional Neural Networks (CNN, or short: convnets), which have been introduced in the early 1990ies \citep{LeCun:1989:Backprop}, and which have demonstrated excellent performance recently \citep{KrizhevskySutskeverHinton:2012:ImagenetDeep}\footnote{This NIPS-paper has 17,770 citations as of 14.12.2017 12:00 CET}. Technically a CNN is structured as a series of different layers, i.e. convolutional layers, pooling layers, fully connected layers and classification layers (e.g. softmax). Convolutional layers are feature maps, where each feature map is connected to local patches in the corresponding feature map in the previous layer. Pooling layers merge similar features into one. Fully connected layers generate the output for the actual task and classification layers produce a label provided a learning instance. CNNs are so popular because the effort of engineering the feature representations can be mostly handled by the network itself.

Also in the medical domain, amazing results have been achieved, e.g. for cancer detection with human-like performance:
\cite{LiuStumpe:2017:GooglePatho} adopted a CNN framework for breast cancer metastasis detection in lymph nodes. By exploiting the information of a pre-trained model, sophisticated image normalization, and building on a multi-stage approach (mimicking the human perception), state-of-the-art methods and even human pathologists have been outperformed on a standard benchmark dataset.

Similarly, \cite{EstevaThrun:2017:DermaNN} addresses the problem of skin cancer detection: they propose a transfer learning setup in which a pre-trained CNN architecture using ImageNet has been modified in such a way that only the final classification layer is discarded and re-trained for the given task (additionally, the parameters are fine-tuned across all layers). Eventually, the obtained network was able to perform on par with human dermatologists on different tasks.

Even though this demonstrates that DL may be beneficial in the medical domain generally, the main challenge is still to cope with the problem that often the rather large amount of training data required is not available. Consequently, there has been a considerable interest in approaches that can learn from a small number of training samples (see section \ref{sect:iML}). The most common and straight forward way is to use data augmentation for constructing iterative optimization via the generation of artificial data or latent variables \citep{VanDykMeng:2001:DataAugmentation,HaubergEtAl:2016:DreamingMoreData}. In our domain image augmentation  \citep{BloiceEtAl:2017:AugmentorMachineLearning} can be particularly beneficial, where additional training samples are generated via variation of the given data, e.g. via rotation, elastic deformation, adding noise, etc.

A prominent example for such an approach is U-Net \citep{RonnebergerEtAl:2015:Unet}, which demonstrated that state-of-the-art results can be obtained for bio-medical image segmentation even when the model was trained on merely a few samples.

Even though this simple approach often yields good results, it is restricted as only limited variations can be generated from the given data. A promising direction is to build on ideas from transfer learning \citep{LongWangJordan:2016:DeepTransfer}.
The key idea is to pre-train a network on large publicly available datasets and then to fine-tune it for the given task. For example, \cite{CaiEtAl:2017:PancreasMRI} fine-tunes the VGG-16 network, which is already pre-trained using a huge amount of natural images, to finally segment pancreas from MR images. In addition, a CRF step is added for the final segmentation. Another way would be to use specific prior knowledge about the actual task \citep{PayerBischof:2016:Landmark}.

However, this information is often not available and, as mentioned above, medical image data and natural images are often not sharing the same characteristics, which is why such approaches often fail in practice.

A totally different way to deal with small amounts of training data is to use synthetically generated samples for training such as \citep{RozantsevEtAl:2015:RenderingImages} which are easy to obtain. However, even in this way the specific characteristics of the given image data might not be reflected properly. To overcome this problem, Generative Adversarial Nets \citep{GoodfellowBengio:2014:adversarialNets} train a generator and a discriminator framework in parallel. The key idea is that the generator synthesizes images and the discriminator decides if an image is real or fake (i.e., generated by the generator). In this way, increasingly better training data can be generated. This idea is for example exploited by \cite{NieEtAl:2017:MedicalImageContext} to better model the nonlinear relationship between CR and MR images.

\subsection{Interpretable Deep Learning Models}
\label{ssect:interpretable_dnn}

Simple models are regarded as more interpretable than complex ones, therefore linear models and basic decision trees still dominate in many applications where interpretability is an issue. This belief is however challenged by recent work, in which carefully designed interpretation techniques have shed light on some of the most complex and deepest machine learning models \citep{SinghRibeiroGuestrin:2016:BlackBoxArxiv}.


Standard supervised convnet models \cite{LeCun:1989:Backprop,KrizhevskySutskeverHinton:2012:ImagenetDeep} map a color 2D input image $x_i$, via a series of layers, to a probability vector $\hat{y_i}$ over $C$ different classes.  Each layer consists of (i) a convolution of the previous layer output (or, in the case of the 1st layer, the input image) with a set of learned filters; (ii) passing the responses through a rectified linear function ({\em $relu(x)=\max(x,0)$}); (iii) [optionally] max pooling over local neighborhoods and (iv) [optionally] a local contrast operation that normalizes the responses across feature maps - refer to \citep{KrizhevskySutskeverHinton:2012:ImagenetDeep} for details. The top few layers of the network are conventional fully-connected networks and the final layer is a softmax classifier \cite{ZeilerFergus:2013:VisDeepArXiV}.

The model can be trained using a large set of $N$ labeled images $\{x,y\}$, where label $y_i$ is a discrete variable indicating the true class. A cross-entropy loss function, suitable for image classification, is used to compare $\hat{y_i}$ and $y_i$. The parameters of the network (filters in the convolutional layers, weight matrices in the fully-connected layers and biases) are trained by back-propagating the derivative of the loss with respect to the parameters throughout the network and updating the parameters via stochastic gradient descent.

Visualizing features in order to gain insight into the behavior of networks has been done previously, however, limited to the first layer(s). In "deeper" layers it is not possible to make projections to the pixel space, consequently there are very few methods to date to provide interpretability of such networks. For example \cite{ErhanBengioCourvilleVincent:2009:TechnicalReportVisDeep} proposed to find the optimal stimulus for each unit by performing gradient descent in image space to maximize each unit's activation. This requires a careful initialization and does not give any information about the unit's invariances. Motivated by this shortcoming, \cite{NigamEtAl:2010:TiledConvolutional} demonstrate how the Hessian of a given unit may be computed numerically around the optimal response, giving some insight into invariances. The problem still remains that for higher layers the invariances are extremely complex and can therefore hardly be captured.

\begin{figure}[htp!]
\vspace{-3mm}
\begin{center}
\includegraphics[width=\textwidth]{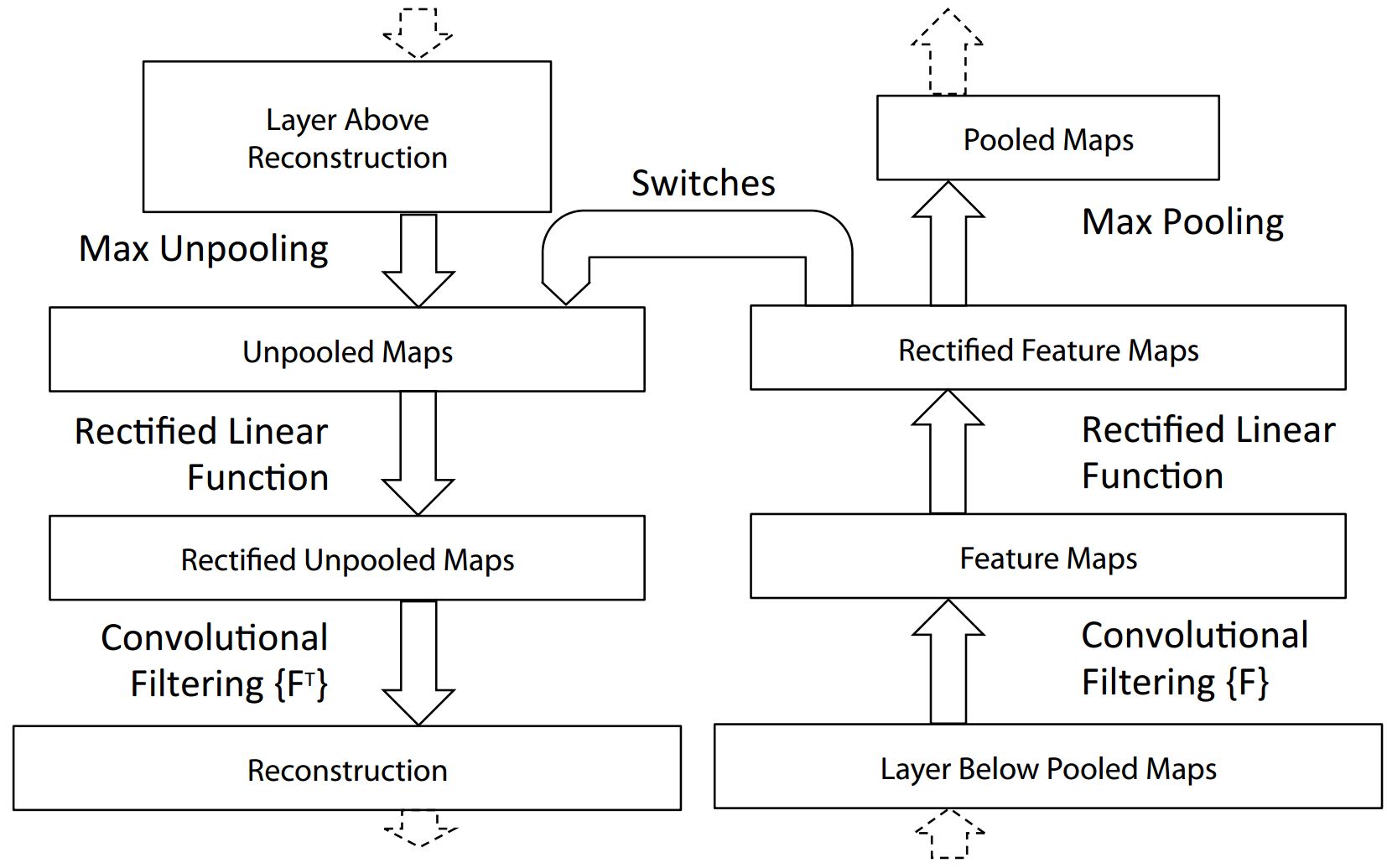}
\end{center}
\vspace*{-0.3cm}
\caption{A deconvnet layer (left) attached to a convnet layer
  (right). The deconvnet is able to reconstruct an approximate version of
  the convnet features from each layer beneath; image source: \cite{ZeilerFergus:2013:VisDeepArXiV}}
\label{fig:deconv}
\vspace*{0.5cm}
\end{figure}

\begin{figure}[htp!]
\vspace{-3mm}
\begin{center}
\includegraphics[width=\textwidth]{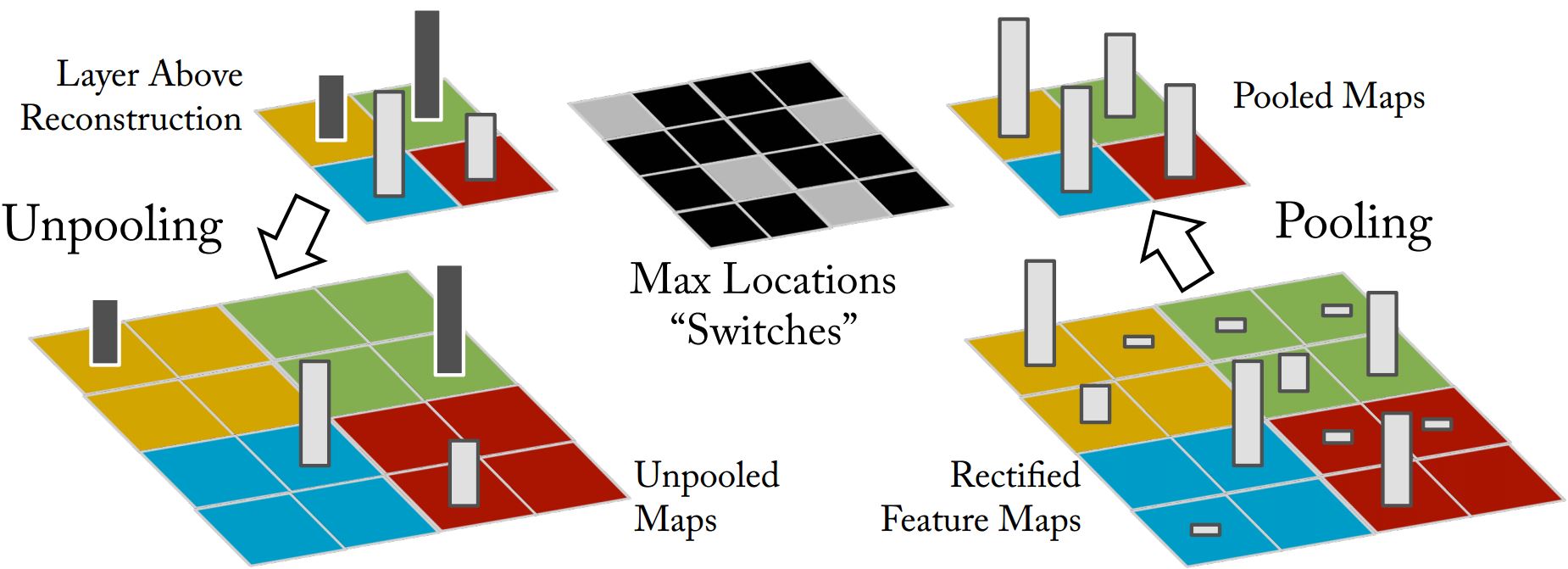}
\end{center}
\vspace*{-0.3cm}
\caption{The unpooling operation in the deconvnet, using {\em switches}
  which record the location of the local max in each pooling region
  (colored zones) during pooling in the convnet; image source: \cite{ZeilerFergus:2013:VisDeepArXiV}}
\label{fig:unpool}
\vspace*{0.1cm}
\end{figure}

A very interesting approach has been developed by \cite{ZeilerFergus:2013:VisDeepArXiV}: they introduce a visualization technique that reveals input stimuli exciting individual feature maps at any layer in the model, which allows to observe a kind of "evolution" of features during the training phase. This allows insight into the internals of the model as well as to discover problems. Their technique is based on a multi-layered deconvolutional network (deconvnet) \citep{ZeilerTaylorFergus:2011:deconvnet} in order to project the feature activations back to the input pixel space. The basic principle can be inferred from Figure~\ref{fig:deconv} and Figure~\ref{fig:unpool}. The procedure is ''unpool, rectify, filter'' in order to reconstruct the activities in the layer which was responsible for an activation. This procedure is repeated until the input pixel space is reached.

\section{Learning from Heterogeneous Graphs}
\label{sect:learning_from_graphs}

Graph Theory \citep{Harary1965StructuralModels,BondyMurty:1976:GraphClassic} provides powerful tools to link data structures of different sources and to reason about probabilistic connections between data objects \citep{Strogatz2001complexNetworks}. Graphical models \citep{Lauritzen:1996:GraphicalModels} combined with probability theory \citep{Kolomogorov:1950:probability}, which can represent a set of random variables and conditional dependencies, form the basis for causality \citep{Pearl:2009:Causality}; this is of extreme importance towards reaching explainable AI. Recent developments in probabilistic programming support such approaches \citep{FadjaRiguzzi:2017:PLP,TranBlei:2017:DeepProbProg}.

An example is the graph-based approach for mitosis extraction in breast cancer from WSI, which has been presented by \cite{RoullierEtAl:2011:GraphsWSI}: Their segmentation uses multi-resolution, which reproduces the WSI examination done by a (human) pathologist. Each resolution level is then analyzed with a focus of attention resulting from a coarser resolution level analysis. At each level a spatial refinement by label regularization is performed to obtain a more accurate segmentation. The proposed segmentation is fully unsupervised by using domain specific knowledge and their strategy for multi-resolution segmentation makes use of regularization on graphs both for image simplification and segmentation. This approach is based on an interesting nonlocal discrete regularization framework on weighted graphs of the arbitrary topologies for image and manifold processing \citep{ElmoatazEtAl:2008:Manifold}, where graphs are the shared representation.

However, the challenge of extracting appropriate graphs out of "natural images" remains, especially w.r.t. identifiable biological structures, which is not an easy task \citep{HolzingerMalleGiuliani:2014:GraphExtraction,HolzingerEtAl2014OnPCD}.

\begin{figure}[!ht]
	\centering
	\includegraphics[width=1\textwidth]{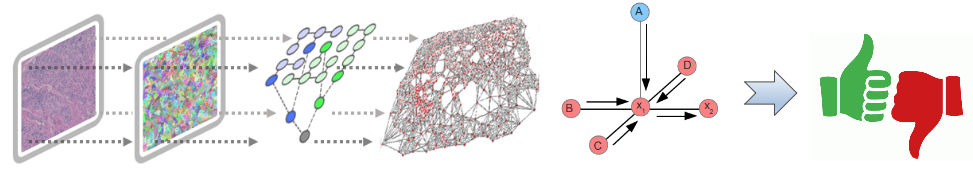}
	\caption{A possible graph computational pipeline: 1) Image preprocessing \& segmentation, 2) projection from pixel to vertex space, 3) graph construction, 4) graph-based data mining, 5) classification}
	\label{fig:pipeline}
\end{figure}

\begin{figure}[ht!]
\includegraphics[width=\textwidth]{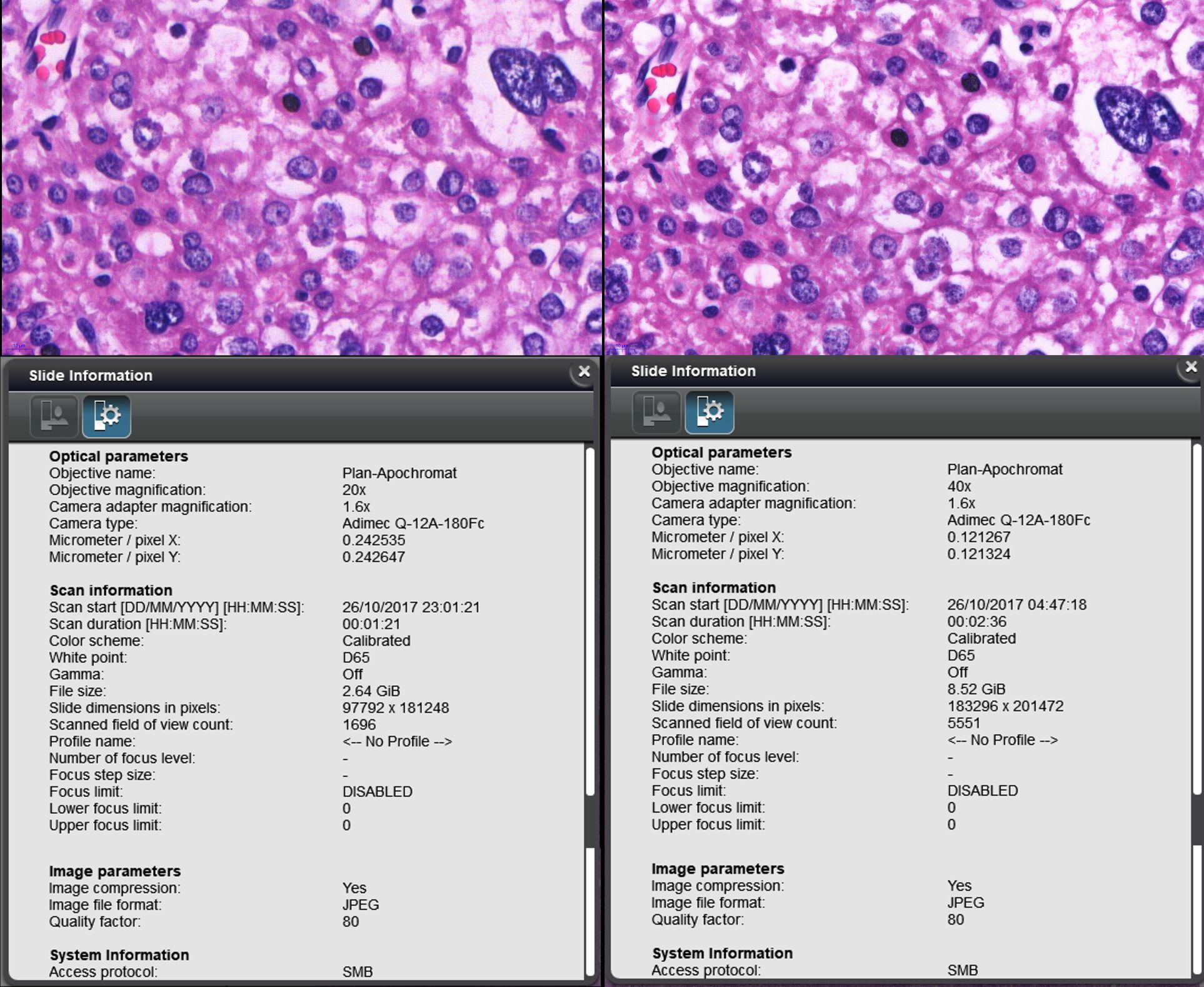}
\caption{FFPE section of a human hepatocellular carcinoma, hematoxylin and eosin staining; comparison of scans with 0,24 micrometer/pixel and 0,12 micrometer/pixel (Image Source: Pathology Graz, obtained with the new P1000 scanner of 3DHistech)}
\label{fig:carcinoma}
\end{figure}

With respect to the sheer size of our pathological input space (see Figure~\ref{fig:carcinoma}) we propose an intriguing possibility of graph-theory: To project an image's pixel space down to a graph's vertex-space, in which each vertex represents a Region-of-Interest (ROI) within the original image (e.g. a cell nucleus). Given a meaningful representation, we can subsequently apply powerful graph algorithms including belief-propagation or Bayesian networks - or even train CNNs using the vertex vector / adjacency matrix as input layer (see Figure~\ref{fig:pipeline}).

While the real power of graph theory does not lie in constructing graphs from individual items but in interlinking heterogeneous sources, simply connecting EHR, *omics and image data of one single patient does not present us with an ''interesting'' graph, since almost no information will be encoded by the graph structure itself (i.e. the topology of the graph), but remain within the ''nodes'' representing the original data. Therefore, it is essential to devise methodologies which allow us to extract features from different data modalities that are \textit{atomic}, \textit{isolated}, \textit{meaningful} and - above all - \textit{linkable} in order to produce a data structure amenable to graph-theoretical procedures. This necessitates the development of a\textit{shared representation on the feature level,} e.g. the mapping of features from different domains to a shared concept space.

While this could be achieved via Word embeddings for EHRs (Word2Vec) and via random walks on image subgraphs / graphs extracted from subimages (Node2Vec), it is yet unclear how to properly perform meaningful embeddings for *omics data. In any case, the goal of such projections is to make data fragments from different sources comparable in terms of structural as well as functional similarity and importance to specific use cases. This could serve as a first step towards explainability, as concept-mappings over modalities would enable text-and-omics-to-screen reasoning algorithms which form part of the idea of a \textit{holistic user interface}.

Finally, merging heterogeneous data sources provides the intriguing possibility of forming superior models about patterns of interplay between symptoms / characteristics on the cellular, *omics, and diagnostic level. Introducing such parameter schemes to generative adversarial networks (GANs) might even allow for the generation of quasi-realistic, synthetic patient data and thus the provision of virtual pathology data sets.

\section{On Topological Data Mining from Medical Images}
\label{sect:topology} \label{sec:topology}

Much potential lies in topological data mining \citep{Holzinger:2014:TopologicalDataMining}, where we need point cloud data sets or distances as inputs. A set of such primitives forms a space, and if we have finite sets equipped with proximity or similarity functions $sim_q \colon S^{q + 1} \to [0, 1]$, which measure how ``close'' or ``similar'' $(q+1)$-tuples of elements of $S$ are, we speak about a \emph{topological space}. A value of $0$ means totally different objects, while $1$ corresponds to equivalent items. This is highly relevant for explainable AI, because natural data form certain structures called manifolds  \citep{AbrahamEtAl:ManifoldsBook,EdelsbrunnerHarerZomrodian:2001:Morse}. Manifolds can be seen as a topological space that is locally homeomorphic (that means it has a continuous function with an inverse function) to a real \textit{n}-dimensional space. In other words: $X$ is a \textit{d}-manifold if every point of $X$ has a neighborhood homeomorphic to $\mathbb{B}^d$; it is called \textit{bounded} if every point has a neighborhood homeomorphic to $\mathbb{B}$ or $\mathbb{B}^{d}_+$ \citep{Cannon:1978:ToplogicalManifold}. This is highly relevant for many applications in machine learning where we are confronted with high-dimensional data and to find meaningful representations in the lower-dimensions \citep{TenenbaumSilvaLangford:2000:Isomap,SaulRoweis:2003:Manifold,WeinbergerSaul:2006:Manifolds}. A topological space can therefor be seen as an abstraction of a metric space, and similarly, manifolds generalize the connectivity of $d$-dimensional Euclidean spaces $\mathbb{B}^{d}$ by being locally similar but globally different. A $d$-dimensional chart at $p \in X$ is a homeomorphism $\phi : U \rightarrow \mathbb{R}^d$  onto an open subset of $\mathbb{R}^d$, where $U$ is a neighborhood of $p$ and open is defined using the metric. A $d$-dimensional manifold ($d$-manifold) is a topological space $X$ with a $d$-dimensional chart at every point $x \in X $ \citep{Zomorodian:2009:COMPTOPinHandbook}.

Also interesting for our work are simplicial complexes ("simplicials") which are spaces described in a very particular way, the basis of which is homology \citep{CerriEtAl:2013:Persistence}. The reason for their existence is that it is not possible to represent surfaces precisely in a computer system due to limited computational storage; thus, surfaces are sampled and represented with triangulations. Such a triangulation is called a simplicial complex, and is a combinatorial space that can represent a space. With such simplicial complexes, the topology of a space can be separated from its geometry.

One way to create a simplicial complex is to examine all subsets of points, and if any subsets of points are close enough, a p-simplex (e.g. line) is added to the complex with those points as vertices. For instance, a Vietoris-Rips complex of diameter $\epsilon$ is defined as $VR(\epsilon)  = {\sigma | diam(\sigma) \leq \epsilon}$, where $diam(\epsilon)$ is defined as the largest distance between two points in $\sigma$. A common way to analyze topological structures is to apply persistent homology \citep{ZomorodianCarlsson:2005:PersistentHomology}, which identifies cluster, holes and voids. It is assumed that more robust topological structures are the one which persist with increasing $\epsilon$.
For us it is very relevant to extract significant features, thus these methods are useful, since they provide robust and general feature definitions with emphasis on global information, e.g. Alpha Shapes \citep{EdelsbrunnerMucke:1994:AlphaShapes}.

The combination of topology and machine learning towards \textit{geometric machine learning} could be very useful in the future, e.g. geometric deep learning, which attempts to generalize structured deep learning models to non-Euclidean domains, such as graphs and manifolds \citep{LeeYoon:2016:TransferLearningDeepGraphs,Bronstein:2017:geometricDeepLearning} and help in developing explainable-AI.

\section{Interactive Machine Learning with the Expert-in-the-Loop}
\label{sect:iML}

The central challenge in machine learning is the development of a model which relates observations $X$ to target variables $Y = {x_1, x_2, x_3, ...}$, inferring either decisions or predictions. While traditionally we deal with deterministic values for $Y$, in some domains - such as pathology - we need to consider uncertainties and therefore seek applications where $Y$ includes unknowns or rare categories - and where the application of automatic machine learning approaches may result in the danger of modeling artifacts.

A further problem is that the complexity of automated machine learning algorithms has generally detained non-experts from the application of such solutions. However, for solving real-world problems, the integration of a domain expert's knowledge can sometimes be indispensable and the interaction of the domain expert with the data would greatly enhance the knowledge discovery process, consequently iML puts the pathologist-in-the-loop to enable what neither a human nor a computer could do on their own \citep{Holzinger:2016:iML}.

One approach to mitigate these shortcomings is to involve humans in the model-building effort; traditionally this has been restricted to having people label data in a pre-processing stage or by requesting experts to select or weight features (bias-injection) before the actual Machine Learning happens. A real \textit{interactive} approach though requires the interplay of algorithmic action and human feedback in a continuous loop, thereby enabling the algorithm to adapt it's internal learning strategy on-the-fly, thus effectively opening up the black box approach \citep{Holzinger:2016:iML,Holzinger:2016:definitionIML} to a glass box\footnote{Our project page is accessible via: https://hci-kdd.org/project/iml}.

This method brings with it two major benefits: 1) for problems in high-dimensional search spaces, it allows the algorithm to restrict possibilities by making use of human experience translated into heuristic functions, and 2) as decisions are always - at least partly - based on inputs stemming from the realm of human expertise, any subsequently automated ML approach will exhibit the same patterns of reasoning, which makes algorithmic decisions much more explainable from a human perspective.

An interesting aspect of such interaction is the Bayesian manner by which humans assign significance to data points they observe and the hypotheses they extract from such data; instead of seeing ''importance of data'' and ''hypothesis'' as separate entities, one should more appropriately model the mutual influence as ''importance of data, given some hypothesis'' and ''hypothesis, given some important data points''.

In combination with \hyperref[sect:learning_from_graphs]{heterogeneous data graphs} and a \textit{holistic} user interface, we can conceive of a progressive interaction cycle where reasoning algorithms present the pathologist with their insights in the form of recommendations (e.g. diagnosis, prognosis, valuable related information) which a pathologist could accept or reject (see Figure \ref{fig:pipeline}, providing continuous feedback for the algorithm to change it's internal mapping parameters. This feedback could be collected implicitly via \textit{affective computing}, i.e. the observation of a user's reaction to content without eliciting conscious, deliberate feedback. This can be done by microscope eye-tracking \citep{Duchowski:2007:eyetracking} which can provide novel insights into the analysis behavior of a pathologist \citep{JaarsmaEtAl:2014:ProcessingHisto,KrupinskiEtAl:2006:EyeMovement}; comparison studies can be enormously useful for the machine learning community\footnote{Compare to the work done by \cite{GopnikEtAl:2004:CausalLearning}}.

Projecting this approach further into the future, we could go beyond gaining insights from moment-to-moment user observation to a system of intention mining, where a chain of user (inter)actions is utilized to derive a pathologist's individual strategy of dealing with a given problem when presented specific observations. The resulting models of individual learning strategies could be analyzed w.r.t. to their historic success, beneficial strategy phases could be extracted from each expert's profile and merged into a \textit{virtual pathologist} combining the best traits of each human archetype. Although the practical deployment of such a ''super pathologist'' might evoke negative reactions from medical experts and patients alike, it would present an ideal solution in education and training of future practitioners (especially given the shortage of expert personnel in the field), and contribute to the research in AI generally and ML specifically.

\section{Explainability of Machine Decisions}
\label{explainability}

Being a rather general term, explainability is usually defined as making automated ML decisions transparent, which one can decompose into meaning \textit{interpretable} (what effect was that decision supposed to have?), \textit{comprehensible} (which data led to the decision?) and \textit{reproducible} (given the data and specific circumstances, can we model a function that outputs the same decision?). In order to achieve this, we need to identify and disentangle explanatory factors in the underlying data, may it be pixels in images, tokens in text, or base-pairs in genetic sequences.


Apart from several references to explainability in earlier sections, a few generic approaches have been recently proposed which in our view warrant specific presentation. In ~\cite{RibeiroSinghGuestrin:2016:Trust} the authors introduce a model they call \textit{LIME} - Local Interpretable Model Explanation - which aims to explain any classification or prediction result by approximating it locally with an interpretable model. They achieve this via sampling around local instances until they arrive at a linear approximation of the global decision function which behaves in a locally faithful way.

A very recent work \cite{BachMuellerEtAl:2015:ETR} describes \textit{layer-wise relevance propagation} (LRP) which the authors claim applicable for any form of layered computation architecture. The method preserves constraints to a total relevance score per layer, so that input dimensions ''compete'' for their influence on the next layer. The system works by first computing the overall function and then working backwards via distributing the relevance of a neuron $k$ at layer $l+1$ onto its input neurons at layer $l$ all the way back to the input dimensions (e.g. pixels within an image). These can then be visualized via heat maps, offering an intuitive interface for human comprehension.

An interesting application framework for explainable interactive ML is \textit{Bonsai \footnote{https://bons.ai}}, which allows programmers without extensive ML experience to train so-called ''curricula'', i.e. bottom-up learning strategies for particular tasks in a Domain Specific Language. These programs are separated into different learning sections and phases, so that the user acts as a constraint-giver to the solution space during training (which has the downside of limiting the universal learning-capabilities of deep architectures), so that mis-classifications of the trained model in the testing phase can be clearly attributed to a ''sub-program'', thereby helping explainability and model acceptance.

\section{Privacy Aware Machine Learning}
\label{sect:paml}

The General Data Protection Regulation (GDPR and ISO/IEC 27001) coming into effect on May 25\ts{th} 2018 attracts a lot of attention in the medical area, as it is concerned with all major issues regarding the processing of personal sensitive information and deals with the legal requirements for data collection, consent regarding processing, anonymization/pseudonymization, data storage, transparency and deletion (\textit{''the right to be forgotten''} \citep{MalleEtAl:2016:forgotten}, calling for federated learning approaches \citep{MalleEtAl:2017:FederatedLearning}).

Still, the major issue is that many details are currently not defined, e.g. whether deletion needs to be done on a physical or simply a logical level, or how strong the anonymization-factors need to be~\citep{VillarongaEtAl:2017:HumansForget}.

Furthermore, some parts are formulated in a way that cannot be achieved with current technological means, e.g. de-anonymization being impossible in any case, as well as the antagonism between deletion and transparency. Thus, the issue of securing sensitive information is one of the big challenges in machine learning in health related environments.

\subsection{Digital Self-Determination}
\label{ssect:deletion}

Providing digital self-determination is one of the major targets of the GDPR and forms the background behind several aspects like (i) the requirement for consent, (ii) no consent given implying the need for anonymization, (iii) transparency of the whole analysis process, including the right of the data subject to promptly receive all information regarding the processing of his/her data, as well as (iv) the right to rectify any erroneous information entered into the processing system, even granting (v) the right to have all personal data deleted. This \textit{''right to be forgotten''} can even be called upon in case explicit consent was granted beforehand, meaning consent can be revoked by the data subject basically at all times.\\

The main problem behind deletion, which is also true for rectification of erroneous data, is that there is currently no definition of the word ''deletion'' within the GDPR(~\cite{VillarongaEtAl:2017:HumansForget}). Furthermore, the GDPR is written as a legal text with a lot of absolutes that can be interpreted as to demand things that are technically impossible like physical deletion in cloud environments. Another issue regarding deletion are backups that must not be changed with respect to backup policies and for security reasons. Also the right to transparency can pose a severe problem, as traceability in complex enrichment processes might require complex Audit \& Control mechanisms that will contain parts of the user data.

While in theory, deleting data from a system is not considered a major problem, in the reality of complex systems \emph{provable deletion}, i.e. being able to guarantee that the deleted data cannot be restored, is far from trivial~ \citep{VillarongaEtAl:2017:HumansForget}. This becomes especially problematic when considering complex systems like databases that on the one hand store most of the data in real-life environments and are therefore the main concern for data deletion and on the other hand have to cater to requirements like ACID-compliance~\citep{HaerderReuter:1983:Database}.

\begin{figure}
\centering
\includegraphics[width=0.95\textwidth]{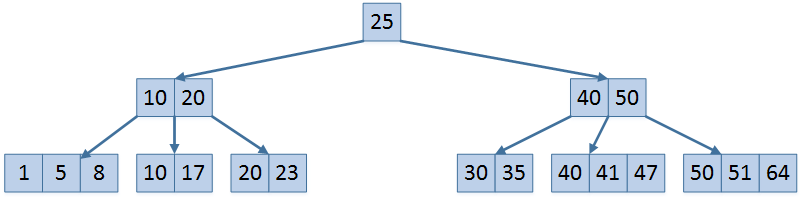}
\caption{Example for a $B^+-Tree$}
\label{fig:btp}
\end{figure}

While manipulating the data requires quite some knowledge of the internal mechanisms of the DBMS in question, in the end it is just a legal question on how much resilience a deletion mechanism has to provide. To further illustrate this aspect, we stay within the database environment and have a look at ''normal'' deletion: Quite like in most file systems, data is not actually deleted from the disk, e.g. by overwriting the memory, but just unlinked from the global search indices: Typically, relational databases store their content inside a special structure called a $B^+$-tree~\citep{BayerMcCreight:2002:Organization} (see Figure~\ref{fig:btp} for an example, still, reality is a bit more complex, but this simplification does not significantly change our point):

\begin{itemize}
\item For the number $m_i$ of elements of node $i$ holds $\frac{d}{2}\leq i \leq d$, where $d$ is a pre-defined value (the order) for the whole tree. Only the root node is allowed to possess less elements ($>0$). Inside the nodes, the elements are stored as sorted lists.
\item Each inner node holding $m$ elements possesses $m+1$ child nodes.
\item The inner nodes of the $B^+$-tree do not store actual information on the elements, but just the information required to let the search engine navigate through the tree efficiently. All actual data is stored in the leafs, which are all on the same level in the tree (balanced tree).
\end{itemize}

\subsection{Unsolicited Propagation of Data}
\label{ssect:data_propagation}

Within a lot of research environments, cooperation with external experts or with academic peers is very important. Within many data driven research approaches this sooner or later leads to the topic of data exchange, e.g. by sharing data within a common environment or providing experts with sample data. While regulations and legal frameworks like the GDPR deal with the topic of protecting the data subjects, another issue is often neglected: The inherent value of even well-anonymized data, or data sets that do not contain sensitive personal information at all. While not a Privacy issue, a lot of data that is usable enough for providing insight on a certain topic does possess an inherent value that needs to be protected, as it is often a vital asset of the (academic/medical) institution. This is especially true for large collections of pathological data that have been obtained through an expensive process of physical labor. Thus, when institutions join cooperative research projects, they must make sure to be able to protect these data assets.\\

Typical measures proposed by traditional security research mainly fall into the category of proactive measures, typically limiting the actual data exchange. Examples for these \emph{proactive measures} include the setup of sealed research environments, typically virtualized environments containing all required data and the analysis tools while not allowing the extraction of detailed information, or splitting the information over several independent processing entities that are assumed to not collude (e.g.~\cite{HudicEtAl:2013:DataConfidentiality}), or providing data only in anonymized or aggregated form.\\

Proactive measures are of course massively interesting but cannot be used under all circumstances, e.g. when the analysts require their own analysis environments or need the data with a certain quality. \emph{Reactive} approaches on the other hand do not aim at limiting unsolicited propagation, but rather allow to detect it, thus allowing the data owner to recoup their losses with legal actions. Fingerprinting/watermarking of data is the most popular approach in this area. Fingerprinted data contains certain changes to the original information that allow the data owner to determine and prove which partner was the source of a data leak. These fingerprints are required to be stable against colluding attackers and it must not be possible to identify and remove the marks from a marked set. Furthermore, they must be unforgeable in order to protect innocent partners against wrongful accusations. In addition, the less information is required to detect the data leak, the better. While many fingerprinting approaches rely on the addition or change of data records, there are also approaches that utilize the intrinsic features of anonymization \citep{KiesebergEtAl:2014:protecting}, which can be very useful in case the data has to be anonymized anyways.

\subsection{Manipulation Detection}
\label{manipulations}

Especially when considering expert-in-the-loop systems in the medical area, the question of securing information against manipulation gains increasing importance. In case experts are used to control and steer the decision making process, their input might be responsible for the well-being of patients, while used in a way that makes it hard for them to actually control the effects, as the data is taken as input by machine learning algorithms and utilized as e.g. knowledge base.

Thus, one vital aspect lies in gaining trust of the doctor in the loop \citep{KiesebergEtAl:2016:TrustDocInLoop,KiesebergEtAl:2016:PrivacyMLDocInLoop} i.e. assuring that the information that was presented as starting point for giving expertise is not changed afterwards. In addition, the expert must be sure that the expertise (the information) gathered from him has not been distorted, i.e. in case the algorithms produce wrong results, the blame cannot be shifted towards him/her without justification. This is especially important in order to protect the experts against cover-ups, e.g. other experts or the developers of the machine learning algorithms trying to put the blame towards them. In our experience, providing such mechanisms is of vital importance in order to attract expert involvement and make them share their knowledge, as well as to increase the acceptance of the results of semi-automated decision engines.\\

While many of the tasks required to establish a secure environment harken back to typical topics of system hardening against inside and outside attacks, i.e. making it as impenetrable as possible, and are therefore highly dependent on the actual technical system in place, the data needs to be secured against another attack vector typically not included in security assessments: The manipulation of the data by the system itself, i.e. typically implemented in the form of a database management system. While the system and its administrators are typically trusted to a certain extent in most architectures, they need to be considered as a major antagonist for data manipulation in order to cover up the application of wrong algorithms or the provisioning of erroneous data. Thus, every piece of information that is used in the analysis needs to be secured against subsequent manipulation attacks, even from users possessing administrator privileges.\\

\begin{figure}
\centering
\includegraphics[width=0.70\textwidth]{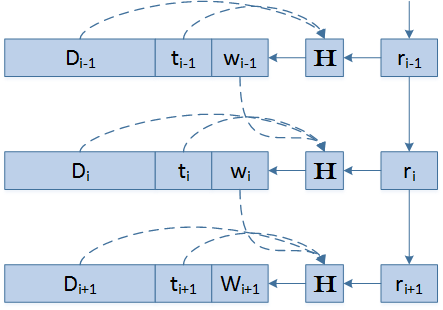}
\caption{Chaining log entries}
\label{fig:chainedwitnesses}
\end{figure}

In addition, other approaches have been introduced including utilization of the database replication mechanism etc., still, research in this area is rather at its infancy and has not received much attention in the past as most real-life systems use traditional system hardening measures together with organizational measures.

\section{Conclusion and Future Outlook}

Digital pathology poses manifold challenges for AI research generally and ML research specifically. Solutions will benefit education, training, research and clinical decision support. The setting of digital pathology is ideally suited to study both human learning and human decision making and contrast it to machine learning and machine decision making. By comparing both the strengths/weaknesses of machine as well as human intelligence it will be possible to find solutions where we are currently lacking appropriate methods.

Future medicine relies increasingly on various data for better detection and treatment of diseases. In order to extract knowledge from massive amounts of information contained in high resolution images, AI and ML methods are indispensable. Moreover, digital pathology provides a fertile environment to integrate data from heterogeneous and distributed sources (images, text, \*omics), making novel information accessible and quantifiable - which is not accessible and quantifiable for the human expert to date. A major challenge in this area will be to ''break down'' the constituent parts of each data source (image region, gene sub-sequence, EHR entry) and to make them interlinkable on a level allowing for reasoning between sensory records, image anomalies \& textual description of symptoms. Eventually, this will enable completely new knowledge and query-based systems, where experts may input an image region and receive genetic sequences potentially explaining the observation.

AI-augmented digital pathology will change the education and training of pathologists, which is an urgently needed solution to the global shortage of medical specialists. In order to realize this potential, we will have to define interaction points between pathologists and the learning system; this entails studies to discover which internal learning states can be best represented in a human-intelligible way, what kind of problems the expert is asked to decide (accept/reject a final recommendation or help identifying image regions), and how to incorporate the human decision back into the internal machine model. Defining a new standard protocol for such interactions would also enable external software providers to hook into the process, generating new business models for interactive diagnostics.

Providing insights into the inner workings of algorithms is not only a future legal requirement but also a valuable tool for ''debugging'', especially in the case of deep neural architectures where exact coefficients at exact moments are to date nearly uninterpretable by human experts. Building on recent works we will in future work experiment with the interplay of iML and explainability - injecting search space constraints based on user decisions gathered from an interaction loop into our programs. Consequently, it is necessary to contrast any gains in explainability with the potential loss of generality incurred by the restriction of algorithmic freedom.

Although deep learning has long proven it's worth in a multitude of ML challenges, we still need to make it efficient on extremely large images (such as WSI). Our main challenge will lie in finding more compact, yet smart image representations e.g. via graph extraction, offering the possibility to ''asymmetrically'' compress an image via region-to-node projection, i.e. applying different compression rates to regions of different importance. As neural nets take simple feature vectors as their inputs, we can easily replace a pixel representation with a node feature vector or the entries of an adjacency / Laplacian matrix. This approach also promises insights above and beyond image classification, since in the process we will experiment with DNNs ''intuiting'' network properties like degree distributions or centralities, potentially yielding deep approximate graph algorithms for problems whose algorithmic runtime has not improved in decades.

As modern AI/ML approaches require great amounts of data, training on local databases could easily hit its limits, yet legal requirements as well as societal expectations restrict us from freely sharing vulnerable data. In order to overcome the unsatisfying trade-off we are confronted with today - lose precision via learning on anonymized data, or risk severe repercussions, we need to examine methods to learn from other models instead of other data.

Eventually, we strive to combine parallel \& distributed DNNs working within their respective ''local sphere'' into a powerful network of learners, each fueled by their unique, private knowledge bases yet capable of incorporating model insights from their peer learners - achieving the grand goal of secure, efficient and reliable privacy aware federated learning.

\subsubsection*{Acknowledgments}

We are grateful for valuable discussions and comments from our local, national and international colleagues. We thank our industrial partner 3DHistech for their assistance with our new scanner and the Biobank Graz for valuable support. Acknowledged Funding: Hochschulraum-Strukturmittelverordnung.

\bibliographystyle{unsrtnat}
\bibliography{references}

\begin{thebibliography}{117}
\providecommand{\natexlab}[1]{#1}
\providecommand{\url}[1]{\texttt{#1}}
\expandafter\ifx\csname urlstyle\endcsname\relax
  \providecommand{\doi}[1]{doi: #1}\else
  \providecommand{\doi}{doi: \begingroup \urlstyle{rm}\Url}\fi

\bibitem[Abraham et~al.(1988)Abraham, Marsden, and
  Ratiu]{AbrahamEtAl:ManifoldsBook}
Ralph Abraham, Jerrold~E. Marsden, and Tudor Ratiu.
\newblock \emph{Manifolds, tensor analysis, and applications}.
\newblock Springer, New York, 1988.
\newblock \doi{10.1007/978-1-4612-1029-0}.

\bibitem[Afework et~al.(1998)Afework, Beynon, Bustamante, Cho, Demarzo,
  Ferreira, Miller, Silberman, Saltz, and
  Sussman]{AfeworkEtAl:1998:Telepathology}
Asmara Afework, Michael~D. Beynon, Fabian Bustamante, Soon Cho, Angelo Demarzo,
  Renato Ferreira, Robert Miller, Mark Silberman, Joel Saltz, and Alan Sussman.
\newblock Digital dynamic telepathology - the virtual microscope.
\newblock In \emph{Proceedings of the AMIA Symposium}, pages 912--916. American
  Medical Informatics Association, 1998.

\bibitem[Al-Janabi et~al.(2012)Al-Janabi, Huisman, and
  Van~Diest]{AlJanabiEtAl:2012:DigitalHistopathology}
Shaimaa Al-Janabi, Andr\'{e} Huisman, and Paul~J. Van~Diest.
\newblock Digital pathology: current status and future perspectives.
\newblock \emph{Histopathology}, 61\penalty0 (1):\penalty0 1--9, 2012.
\newblock \doi{10.1111/j.1365-2559.2011.03814.x}.

\bibitem[Arel et~al.(2010)Arel, Rose, and
  Karnowski]{ArelRoseKarnowski:2010:DLoverview}
Itamar Arel, Derek~C. Rose, and Thomas~P. Karnowski.
\newblock Deep machine learning - a new frontier in artificial intelligence
  research [research frontier].
\newblock \emph{IEEE Computational Intelligence Magazine}, 5\penalty0
  (4):\penalty0 13--18, 2010.
\newblock \doi{10.1109/MCI.2010.938364}.

\bibitem[Arteta et~al.(2012)Arteta, Lempitsky, Noble, and
  Zisserman]{ArtetaZisserman:2012:Celldetection}
Carlos Arteta, Victor Lempitsky, J.~Alison Noble, and Andrew Zisserman.
\newblock Learning to detect cells using non-overlapping extremal regions.
\newblock In \emph{International Conference on Medical Image Computing and
  Computer-Assisted Intervention MICCAI}, pages 348--356. Springer, Heidelberg,
  2012.
\newblock \doi{10.1007/978-3-642-33415-3_43}.

\bibitem[Asada and Doi(1997)]{AsadaDoi:1997:PatentDD}
Naoki Asada and Kunio Doi.
\newblock Method and system for differential diagnosis based on clinical and
  radiological information using artificial neural networks.
\newblock \emph{Patent US5622171 A}, 1997.

\bibitem[Bach et~al.(2015)Bach, Binder, Montavon, Klauschen, M\"{u}ller, and
  Samek]{BachMuellerEtAl:2015:ETR}
Sebastian Bach, Alexander Binder, Gr\'{e}goire Montavon, Frederick Klauschen,
  Klaus-Robert M\"{u}ller, and Wojciech Samek.
\newblock On pixel-wise explanations for non-linear classifier decisions by
  layer-wise relevance propagation.
\newblock \emph{PloS one}, 10\penalty0 (7):\penalty0 e0130140, 2015.
\newblock \doi{10.1371/journal.pone.0130140}.

\bibitem[Barbareschi et~al.(2000)Barbareschi, Demichelis, Forti, and
  Palma]{BarbareschiEtAl:2000:DigitalPathology}
Mattia Barbareschi, Francesca Demichelis, Stefano Forti, and Paolo~Dalla Palma.
\newblock Digital pathology: Science fiction?
\newblock \emph{International journal of surgical pathology}, 8\penalty0
  (4):\penalty0 261--263, 2000.

\bibitem[Barisoni et~al.(2017)Barisoni, Gimpel, Kain, Laurinavicius, Bueno,
  Zeng, Liu, Schaefer, Kretzler, and
  Holzman]{BarisoniEtAl:2017:Standardization}
Laura Barisoni, Charlotte Gimpel, Renate Kain, Arvydas Laurinavicius, Gloria
  Bueno, Caihong Zeng, Zhihong Liu, Franz Schaefer, Matthias Kretzler, and
  Lawrence~B. Holzman.
\newblock Digital pathology imaging as a novel platform for standardization and
  globalization of quantitative nephropathology.
\newblock \emph{Clinical Kidney Journal}, 10\penalty0 (2):\penalty0 176--187,
  2017.
\newblock \doi{10.1093/ckj/sfw129}.

\bibitem[Bayer and McCreight(2002)]{BayerMcCreight:2002:Organization}
Rudolf Bayer and Edward McCreight.
\newblock Organization and maintenance of large ordered indexes.
\newblock In \emph{Software pioneers}, pages 245--262. Springer, 2002.
\newblock \doi{10.1007/978-3-642-59412-0_15}.

\bibitem[Blanchet and Smolinska(2016)]{Blanchet:2016:DataFusionMetabolomics}
Lionel Blanchet and Agnieszka Smolinska.
\newblock Data fusion in metabolomics and proteomics for biomarker discovery.
\newblock In Klaus Jung, editor, \emph{Statistical Analysis in Proteomics},
  pages 209--223. Springer, New York, 2016.
\newblock \doi{10.1007/978-1-4939-3106-4_14}.

\bibitem[Bloice et~al.(2017)Bloice, Stocker, and
  Holzinger]{BloiceEtAl:2017:AugmentorMachineLearning}
Marcus~D Bloice, Christof Stocker, and Andreas Holzinger.
\newblock Augmentor: An image augmentation library for machine learning.
\newblock \emph{arXiv preprint arXiv:1708.04680}, 2017.

\bibitem[Bondy and Murty(1976)]{BondyMurty:1976:GraphClassic}
John~A. Bondy and Uppaluri S.~R. Murty.
\newblock \emph{Graph theory with applications}.
\newblock Elsevier, New York, Amsterdam, Oxford, 1976.

\bibitem[Bronstein et~al.(2017)Bronstein, Bruna, LeCun, Szlam, and
  Vandergheynst]{Bronstein:2017:geometricDeepLearning}
Michael~M. Bronstein, Joan Bruna, Yann LeCun, Arthur Szlam, and Pierre
  Vandergheynst.
\newblock Geometric deep learning: Going beyond euclidean data.
\newblock \emph{IEEE Signal Processing Magazine}, 34\penalty0 (4):\penalty0
  18--42, 2017.
\newblock \doi{10.1109/MSP.2017.2693418}.

\bibitem[Cai et~al.(2017)Cai, Lu, Xie, Xing, and
  Yang]{CaiEtAl:2017:PancreasMRI}
Jinzheng Cai, Le~Lu, Yuanpu Xie, Fuyong Xing, and Lin Yang.
\newblock Pancreas segmentation in mri using graph-based decision fusion on
  convolutional neural networks.
\newblock In \emph{International Conference on Medical Image Computing and
  Computer-Assisted Intervention (MICCAI), Lecture Notes in Computer Science
  LNCS 10435}, pages 674--682. Springer, Cham, 2017.
\newblock \doi{10.1007/978-3-319-66179-7_77}.

\bibitem[Cannon(1978)]{Cannon:1978:ToplogicalManifold}
James~W. Cannon.
\newblock The recognition problem: what is a topological manifold?
\newblock \emph{Bulletin of the American Mathematical Society}, 84\penalty0
  (5):\penalty0 832--866, 1978.

\bibitem[Cerri et~al.(2013)Cerri, Fabio, Ferri, Frosini, and
  Landi]{CerriEtAl:2013:Persistence}
Andrea Cerri, Barbara~Di Fabio, Massimo Ferri, Patrizio Frosini, and Claudia
  Landi.
\newblock Betti numbers in multidimensional persistent homology are stable
  functions.
\newblock \emph{Mathematical Methods in the Applied Sciences}, 36\penalty0
  (12):\penalty0 1543--1557, 2013.
\newblock \doi{10.1002/mma.2704}.

\bibitem[Dee(2009)]{Dee:2009:VirtualMicroscopyEducation}
Fred~R. Dee.
\newblock Virtual microscopy in pathology education.
\newblock \emph{Human pathology}, 40\penalty0 (8):\penalty0 1112--1121, 2009.

\bibitem[Duchowski(2007)]{Duchowski:2007:eyetracking}
Andrew~T. Duchowski.
\newblock \emph{Eye tracking methodology: Theory and practice}.
\newblock Springer, Cham, 2007.
\newblock \doi{10.1007/978-3-319-57883-5}.

\bibitem[Edelsbrunner and M\"{u}cke(1994)]{EdelsbrunnerMucke:1994:AlphaShapes}
Herbert Edelsbrunner and Ernst~P. M\"{u}cke.
\newblock 3-dimensional alpha-shapes.
\newblock \emph{ACM Transactions on Graphics}, 13\penalty0 (1):\penalty0
  43--72, 1994.
\newblock \doi{10.1145/174462.156635}.

\bibitem[Edelsbrunner et~al.(2001)Edelsbrunner, Harer, and
  Zomorodian]{EdelsbrunnerHarerZomrodian:2001:Morse}
Herbert Edelsbrunner, John Harer, and Afra Zomorodian.
\newblock Hierarchical morse complexes for piecewise linear 2-manifolds.
\newblock In \emph{Proceedings of the seventeenth annual symposium on
  Computational geometry}, pages 70--79. ACM, 2001.
\newblock \doi{10.1145/378583.378626}.

\bibitem[Elmoataz et~al.(2008)Elmoataz, L\'{e}zoray, and
  Bougleux]{ElmoatazEtAl:2008:Manifold}
Abderrahim Elmoataz, Olivier L\'{e}zoray, and S\'{e}bastien Bougleux.
\newblock Nonlocal discrete regularization on weighted graphs: A framework for
  image and manifold processing.
\newblock \emph{IEEE Transactions on Image Processing}, 17\penalty0
  (7):\penalty0 1047--1060, 2008.
\newblock \doi{10.1109/TIP.2008.924284}.

\bibitem[Engelbart(1995)]{Engelbart:1995:Augmenting}
Douglas~C. Engelbart.
\newblock Toward augmenting the human intellect and boosting our collective
  {IQ}.
\newblock \emph{Communications of the ACM}, 38\penalty0 (8):\penalty0 30--32,
  1995.
\newblock \doi{10.1145/208344.208352}.

\bibitem[Engelbart and English(1968)]{EngelbartEnglish:1968:StanfordCenter}
Douglas~C. Engelbart and William~K. English.
\newblock A research center for augmenting human intellect.
\newblock In \emph{AFIPS 68, Fall joint computer conference}, pages 395--410.
  ACM, 1968.
\newblock \doi{10.1145/1476589.1476645}.

\bibitem[Epstein et~al.(1996)Epstein, Walsh, and
  Sanfilippo]{EpsteinEtAL:1996:SecondOpinion}
Jonathan~I. Epstein, Patrick~C. Walsh, and Fred Sanfilippo.
\newblock Clinical and cost impact of second-opinion pathology: Review of
  prostate biopsies prior to radical prostatectomy.
\newblock \emph{The American Journal of Surgical Pathology}, 20\penalty0
  (7):\penalty0 851--857, 1996.

\bibitem[Erhan et~al.(2009)Erhan, Bengio, Courville, and
  Vincent]{ErhanBengioCourvilleVincent:2009:TechnicalReportVisDeep}
Dumitru Erhan, Yoshua Bengio, Aaron Courville, and Pascal Vincent.
\newblock Visualizing higher-layer features of a deep network.
\newblock \emph{University of Montreal Technical Report Nr. 1341}, 2009.

\bibitem[Esteva et~al.(2017)Esteva, Kuprel, Novoa, Ko, Swetter, Blau, and
  Thrun]{EstevaThrun:2017:DermaNN}
Andre Esteva, Brett Kuprel, Roberto~A. Novoa, Justin Ko, Susan~M. Swetter,
  Helen~M. Blau, and Sebastian Thrun.
\newblock Dermatologist-level classification of skin cancer with deep neural
  networks.
\newblock \emph{Nature}, 542\penalty0 (7639):\penalty0 115--118, 2017.
\newblock \doi{10.1038/nature21056}.

\bibitem[Fadja and Riguzzi(2017)]{FadjaRiguzzi:2017:PLP}
Arnaud~N. Fadja and Fabrizio Riguzzi.
\newblock Probabilistic logic programming in action.
\newblock In Andreas Holzinger, Randy Goebel, Massimo Ferri, and Vasile Palade,
  editors, \emph{Towards Integrative Machine Learning and Knowledge Extraction:
  BIRS Workshop, Banff, AB, Canada, July 24-26, 2015, Revised Selected Papers},
  pages 89--116. Springer, Cham, 2017.
\newblock \doi{10.1007/978-3-319-69775-8_5}.

\bibitem[Gell et~al.(2000)Gell, Madjaric, Leodolter, K\"{o}le, and
  Leitner]{GellEtAl:2000:HospitalInformationSystem}
G\"{u}nther Gell, Miroslav Madjaric, Werner Leodolter, Wolfgang K\"{o}le, and
  Hubert Leitner.
\newblock {HIS} purchase projects in public hospitals of styria, austria.
\newblock \emph{International Journal of Medical Informatics}, 58-59:\penalty0
  147--155, 2000.
\newblock \doi{10.1016/S1386-5056(00)00083-6}.

\bibitem[Goodfellow et~al.(2014)Goodfellow, Pouget-Abadie, Mirza, Xu,
  Warde-Farley, Ozair, Courville, and
  Bengio]{GoodfellowBengio:2014:adversarialNets}
Ian Goodfellow, Jean Pouget-Abadie, Mehdi Mirza, Bing Xu, David Warde-Farley,
  Sherjil Ozair, Aaron Courville, and Yoshua Bengio.
\newblock Generative adversarial nets.
\newblock In Zhoubin Ghahramani, Max Welling, Corinna Cortes, Neil~D. Lawrence,
  and Kilian~Q. Weinberger, editors, \emph{Advances in neural information
  processing systems (NIPS)}, pages 2672--2680, 2014.

\bibitem[Goodfellow et~al.(2016)Goodfellow, Bengio, and
  Courville]{GoodfellowBengioCourville:2016:DeepLearningBook}
Ian Goodfellow, Yoshua Bengio, and Aaron Courville.
\newblock \emph{Deep Learning}.
\newblock MIT Press, Cambridge (MA), 2016.

\bibitem[Gopnik and et~al.(2004)]{GopnikEtAl:2004:CausalLearning}
Alison Gopnik and et~al.
\newblock A theory of causal learning in children: causal maps and {B}ayes
  nets.
\newblock \emph{Psychological review}, 111\penalty0 (1):\penalty0 3--32, 2004.
\newblock \doi{10.1037/0033-295X.111.1.3}.

\bibitem[Greenspan et~al.(2016)Greenspan, van Ginneken, and
  Summers]{Greenspan:2016:DeepMedicalImaging}
Hayit Greenspan, Bram van Ginneken, and Ronald~M. Summers.
\newblock Guest editorial deep learning in medical imaging: Overview and future
  promise of an exciting new technique.
\newblock \emph{IEEE Transactions on Medical Imaging}, 35\penalty0
  (5):\penalty0 1153--1159, 2016.
\newblock \doi{10.1109/TMI.2016.2553401}.

\bibitem[Haerder and Reuter(1983)]{HaerderReuter:1983:Database}
Theo Haerder and Andreas Reuter.
\newblock Principles of transaction-oriented database recovery.
\newblock \emph{ACM Computing Surveys (CSUR)}, 15\penalty0 (4):\penalty0
  287--317, 1983.
\newblock \doi{10.1145/289.291}.

\bibitem[Hainaut et~al.(2017)Hainaut, Vaught, Zatloukal, and
  Pasterk]{HainautPasterk:2017:Biobanking}
Pierre Hainaut, Jim Vaught, Kurt Zatloukal, and Markus Pasterk.
\newblock \emph{Biobanking of Human Biospecimens}.
\newblock Springer, Cham, 2017.
\newblock \doi{10.1007/978-3-319-55120-3}.

\bibitem[Harary et~al.(1965)Harary, Norman, and
  Cartwright]{Harary1965StructuralModels}
Frank Harary, Robert~Z. Norman, and Dorwin Cartwright.
\newblock \emph{Structural Models: An Introduction to the Theory of Directed
  Graphs}.
\newblock Wiley, New York, 1965.

\bibitem[Hauberg et~al.(2016)Hauberg, Freifeld, Larsen, Fisher, and
  Hansen]{HaubergEtAl:2016:DreamingMoreData}
S{\o}ren Hauberg, Oren Freifeld, Anders Boesen~Lindbo Larsen, John Fisher, and
  Lars Hansen.
\newblock Dreaming more data: Class-dependent distributions over
  diffeomorphisms for learned data augmentation.
\newblock In \emph{19th International Conference on Artificial Intelligence and
  Statistics (AISTATS)}, pages 342--350. JMLR, 2016.

\bibitem[Holzinger(2012)]{Holzinger:2012:DATAconf}
Andreas Holzinger.
\newblock {On Knowledge Discovery and Interactive Intelligent Visualization of
  Biomedical Data - Challenges in Human–-Computer Interaction \& Biomedical
  Informatics}.
\newblock In Markus Helfert, Chiara Fancalanci, and Joaquim Filipe, editors,
  \emph{DATA 2012, International Conference on Data Technologies and
  Applications}, pages 5--16. 2012.

\bibitem[Holzinger(2013)]{Holzinger:2013:HCI-KDD}
Andreas Holzinger.
\newblock {Human–-Computer Interaction and Knowledge Discovery (HCI-KDD):
  What is the benefit of bringing those two fields to work together?}
\newblock In Alfredo Cuzzocrea, Christian Kittl, Dimitris~E. Simos, Edgar
  Weippl, and Lida Xu, editors, \emph{Multidisciplinary Research and Practice
  for Information Systems, Springer Lecture Notes in Computer Science LNCS
  8127}, pages 319--328. Springer, Heidelberg, Berlin, New York, 2013.
\newblock \doi{10.1007/978-3-642-40511-2_22}.

\bibitem[Holzinger(2014{\natexlab{a}})]{Holzinger:2014:SpringerTextbook}
Andreas Holzinger.
\newblock \emph{Biomedical Informatics: Discovering Knowledge in Big Data}.
\newblock Springer, New York, 2014{\natexlab{a}}.
\newblock \doi{10.1007/978-3-319-04528-3}.

\bibitem[Holzinger(2014{\natexlab{b}})]{Holzinger:2014:TopologicalDataMining}
Andreas Holzinger.
\newblock On topological data mining.
\newblock In \emph{Interactive Knowledge Discovery and Data Mining in
  Biomedical Informatics, Lecture Notes in Computer Science LNCS 8401}, pages
  331--356. Springer, Heidelberg, 2014{\natexlab{b}}.
\newblock \doi{10.1007/978-3-662-43968-5_19}.

\bibitem[Holzinger(2014{\natexlab{c}})]{Holzinger:2014:trends}
Andreas Holzinger.
\newblock Trends in interactive knowledge discovery for personalized medicine:
  Cognitive science meets machine learning.
\newblock \emph{{IEEE} Intelligent Informatics Bulletin}, 15\penalty0
  (1):\penalty0 6--14, 2014{\natexlab{c}}.

\bibitem[Holzinger(2016{\natexlab{a}})]{Holzinger:2016:definitionIML}
Andreas Holzinger.
\newblock Interactive machine learning ({iML}).
\newblock \emph{Informatik Spektrum}, 39\penalty0 (1):\penalty0 64--68,
  2016{\natexlab{a}}.
\newblock \doi{10.1007/s00287-015-0941-6}.

\bibitem[Holzinger(2016{\natexlab{b}})]{Holzinger:2016:iML}
Andreas Holzinger.
\newblock Interactive machine learning for health informatics: When do we need
  the human-in-the-loop?
\newblock \emph{Brain Informatics}, 3\penalty0 (2):\penalty0 119--131,
  2016{\natexlab{b}}.
\newblock \doi{10.1007/s40708-016-0042-6}.

\bibitem[Holzinger(2017)]{Holzinger:2017:InauguralMAKE}
Andreas Holzinger.
\newblock Introduction to machine learning and knowledge extraction {(MAKE)}.
\newblock \emph{Machine Learning and Knowledge Extraction}, 1\penalty0
  (1):\penalty0 1--20, 2017.
\newblock \doi{10.3390/make1010001}.

\bibitem[Holzinger et~al.(2000)Holzinger, Kainz, Gell, Brunold, and
  Maurer]{HolzingerEtAl:2000:IntelligentTutoring}
Andreas Holzinger, Andreas Kainz, G\"{u}nther Gell, Max Brunold, and Hermann
  Maurer.
\newblock Interactive computer assisted formulation of retrieval requests for a
  medical information system using an intelligent tutoring system.
\newblock In Jacqueline Bourdeau and Rachelle Heller, editors, \emph{World
  Conference on Educational Media and Technology}, pages 466--471. Association
  for the Advancement of Computing in Education (AACE), 2000.
\newblock URL \url{https://www.learntechlib.org/p/16109}.

\bibitem[Holzinger et~al.(2014{\natexlab{a}})Holzinger, Dehmer, and
  Jurisica]{HolzingerEtAl:2014:KDDBio}
Andreas Holzinger, Matthias Dehmer, and Igor Jurisica.
\newblock Knowledge discovery and interactive data mining in bioinformatics -
  state-of-the-art, future challenges and research directions.
\newblock \emph{{BMC Bioinformatics}}, 15\penalty0 (S6):\penalty0 I1,
  2014{\natexlab{a}}.
\newblock \doi{doi:10.1186/1471-2105-15-S6-I1}.

\bibitem[Holzinger et~al.(2014{\natexlab{b}})Holzinger, Malle, Bloice, Wiltgen,
  Ferri, Stanganelli, and Hofmann-Wellenhof]{HolzingerEtAl2014OnPCD}
Andreas Holzinger, Bernd Malle, Marcus Bloice, Marco Wiltgen, Massimo Ferri,
  Ignazio Stanganelli, and Rainer Hofmann-Wellenhof.
\newblock On the generation of point cloud data sets: Step one in the knowledge
  discovery process.
\newblock In \emph{Interactive Knowledge Discovery and Data Mining in
  Biomedical Informatics, Lecture Notes in Computer Science, LNCS 8401}, volume
  8401, pages 57--80. Springer, Berlin Heidelberg, 2014{\natexlab{b}}.
\newblock \doi{10.1007/978-3-662-43968-5_4}.

\bibitem[Holzinger et~al.(2014{\natexlab{c}})Holzinger, Malle, and
  Giuliani]{HolzingerMalleGiuliani:2014:GraphExtraction}
Andreas Holzinger, Bernd Malle, and Nicola Giuliani.
\newblock On graph extraction from image data.
\newblock In Dominik Slezak, James~F. Peters, Ah-Hwee Tan, and Lars Schwabe,
  editors, \emph{Brain Informatics and Health, BIH 2014, Lecture Notes in
  Artificial Intelligence, LNAI 8609}, pages 552--563. Springer, Heidelberg,
  Berlin, 2014{\natexlab{c}}.
\newblock \doi{10.1007/978-3-319-09891-3_50}.

\bibitem[Holzinger et~al.(2014{\natexlab{d}})Holzinger, Stocker, and
  Dehmer]{HolzingerDehmer:2014:TaxonomyData}
Andreas Holzinger, Christof Stocker, and Matthias Dehmer.
\newblock Big complex biomedical data: Towards a taxonomy of data.
\newblock In Mohammad~S. Obaidat and Joaquim Filipe, editors,
  \emph{Communications in Computer and Information Science CCIS 455}, pages
  3--18. Springer, Berlin Heidelberg, 2014{\natexlab{d}}.
\newblock \doi{10.1007/978-3-662-44791-8_1}.

\bibitem[Holzinger et~al.(2017{\natexlab{a}})Holzinger, Malle, Kieseberg, Roth,
  M\"{u}ller, Reihs, and
  Zatloukal]{HolzingerEtAl:2017:DigitalPathologyMachineLearning}
Andreas Holzinger, Bernd Malle, Peter Kieseberg, Peter~M. Roth, Heimo
  M\"{u}ller, Robert Reihs, and Kurt Zatloukal.
\newblock Machine learning and knowledge extraction in digital pathology needs
  an integrative approach.
\newblock In \emph{Towards Integrative Machine Learning and Knowledge
  Extraction, Springer Lecture Notes in Artificial Intelligence Volume LNAI
  10344}, pages 13--50. Springer, Cham, 2017{\natexlab{a}}.
\newblock \doi{10.1007/978-3-319-69775-8_2}.

\bibitem[Holzinger et~al.(2017{\natexlab{b}})Holzinger, Plass, Holzinger,
  Crisan, Pintea, and Palade]{HolzingerEtAl:2017:glassbox}
Andreas Holzinger, Markus Plass, Katharina Holzinger, Gloria~Cerasela Crisan,
  Camelia-M. Pintea, and Vasile Palade.
\newblock A glass-box interactive machine learning approach for solving np-hard
  problems with the human-in-the-loop.
\newblock \emph{arXiv:1708.01104}, 2017{\natexlab{b}}.

\bibitem[Hudic et~al.(2013)Hudic, Islam, Kieseberg, Rennert, and
  Weippl]{HudicEtAl:2013:DataConfidentiality}
Aleksandar Hudic, Shareeful Islam, Peter Kieseberg, Sylvi Rennert, and Edgar
  Weippl.
\newblock Data confidentiality using fragmentation in cloud computing.
\newblock \emph{International Journal of Pervasive Computing and
  Communications}, 9\penalty0 (1):\penalty0 37--51, 2013.
\newblock \doi{10.1108/17427371311315743}.

\bibitem[Huisman et~al.(2010)Huisman, Looijen, van~den Brink, and van
  Diest]{HuismanEtAl:2010:DigitalPathology}
Andr\'{e} Huisman, Arnoud Looijen, Steven~M. van~den Brink, and Paul~J. van
  Diest.
\newblock Creation of a fully digital pathology slide archive by high-volume
  tissue slide scanning.
\newblock \emph{Human pathology}, 41\penalty0 (5):\penalty0 751--757, 2010.
\newblock \doi{10.1016/j.humpath.2009.08.026}.

\bibitem[Huppertz and Holzinger(2014)]{Huppertz:2014:Biobank}
Berthold Huppertz and Andreas Holzinger.
\newblock Biobanks – a source of large biological data sets: Open problems
  and future challenges.
\newblock In Andreas Holzinger and Igor Jurisica, editors, \emph{Interactive
  Knowledge Discovery and Data Mining in Biomedical Informatics, Lecture Notes
  in Computer Science LNCS 8401}, pages 317--330. Springer, Berlin, Heidelberg,
  2014.
\newblock \doi{10.1007/978-3-662-43968-5_18}.

\bibitem[Jaarsma et~al.(2014)Jaarsma, Jarodzka, Nap, Merrienboer, and
  Boshuizen]{JaarsmaEtAl:2014:ProcessingHisto}
Thomas Jaarsma, Halszka Jarodzka, Marius Nap, Jeroen~J. Merrienboer, and Henny
  Boshuizen.
\newblock Expertise under the microscope: processing histopathological slides.
\newblock \emph{Medical education}, 48\penalty0 (3):\penalty0 292--300, 2014.
\newblock \doi{10.1111/medu.12385}.

\bibitem[Jeanquartier et~al.(2016)Jeanquartier, Jean-Quartier, Schreck,
  Cemernek, and Holzinger]{JeanquartierEtAl:2016:OpenData}
Fleur Jeanquartier, Claire Jean-Quartier, Tobias Schreck, David Cemernek, and
  Andreas Holzinger.
\newblock Integrating open data on cancer in support to tumor growth analysis.
\newblock In Elena~M. Renda, Miroslav Bursa, Andreas Holzinger, and Sami Khuri,
  editors, \emph{Lecture Notes in Computer Science LNCS 9832}, pages 49--66.
  Springer, Cham, 2016.
\newblock \doi{10.1007/978-3-319-43949-5_4}.

\bibitem[Kell(2004)]{Kell:2004:Sensemaking}
Douglas~B. Kell.
\newblock Metabolomics and systems biology: making sense of the soup.
\newblock \emph{Current opinion in microbiology}, 7\penalty0 (3):\penalty0
  296--307, 2004.
\newblock \doi{10.1016/j.mib.2004.04.012}.

\bibitem[Kieseberg et~al.(2014)Kieseberg, Hobel, Schrittwieser, Weippl, and
  Holzinger]{KiesebergEtAl:2014:protecting}
Peter Kieseberg, Heidelinde Hobel, Sebastian Schrittwieser, Edgar Weippl, and
  Andreas Holzinger.
\newblock Protecting anonymity in data-driven biomedical science.
\newblock In Andreas Holzinger and Igor Jurisica, editors, \emph{Interactive
  Knowledge Discovery and Data Mining in Biomedical Informatics, Lecture Notes
  in Computer Science, LNCS 8401}, pages 301--316. Springer, Berlin Heidelberg,
  2014.
\newblock \doi{10.1007/978-3-662-43968-5_17}.

\bibitem[Kieseberg et~al.(2016{\natexlab{a}})Kieseberg, Malle, Fr\"{u}hwirt,
  Weippl, and Holzinger]{KiesebergEtAl:2016:PrivacyMLDocInLoop}
Peter Kieseberg, Bernd Malle, Peter Fr\"{u}hwirt, Edgar Weippl, and Andreas
  Holzinger.
\newblock A tamper-proof audit and control system for the doctor in the loop.
\newblock \emph{Brain Informatics}, 3\penalty0 (4):\penalty0 269–279,
  2016{\natexlab{a}}.
\newblock \doi{10.1007/s40708-016-0046-2}.

\bibitem[Kieseberg et~al.(2016{\natexlab{b}})Kieseberg, Weippl, and
  Holzinger]{KiesebergEtAl:2016:TrustDocInLoop}
Peter Kieseberg, Edgar Weippl, and Andreas Holzinger.
\newblock Trust for the doctor-in-the-loop.
\newblock \emph{European Research Consortium for Informatics and Mathematics
  (ERCIM) News: Tackling Big Data in the Life Sciences}, 104\penalty0
  (1):\penalty0 32--33, 2016{\natexlab{b}}.

\bibitem[Kircher et~al.(2014)Kircher, Larsson, and
  Hultgren]{KircherEtAl:2014:levelsAutomation}
Katja Kircher, Annika Larsson, and Jonas~Andersson Hultgren.
\newblock Tactical driving behavior with different levels of automation.
\newblock \emph{IEEE Transactions on Intelligent Transportation Systems},
  15\penalty0 (1):\penalty0 158--167, 2014.
\newblock \doi{10.1109/TITS.2013.2277725}.

\bibitem[Kobayashi et~al.(1974)Kobayashi, Takatani, Hattori, and
  Kimura]{Kobayashi:1974:differentialDiagnosis}
Toshiji Kobayashi, Osamu Takatani, Nobu Hattori, and Kiyoji Kimura.
\newblock Differential diagnosis of breast tumors. the sensitivity graded
  method of ultrasonotomography and clinical evaluation of its diagnostic
  accuracy.
\newblock \emph{Cancer}, 33\penalty0 (4):\penalty0 940--951, 1974.

\bibitem[Kolmogorov(1950)]{Kolomogorov:1950:probability}
Andre\v{i}~N. Kolmogorov.
\newblock \emph{Foundations of the Theory of Probability}.
\newblock Chelsea, Oxford, 1950.

\bibitem[Krizhevsky et~al.(2012)Krizhevsky, Sutskever, and
  Hinton]{KrizhevskySutskeverHinton:2012:ImagenetDeep}
Alex Krizhevsky, Ilya Sutskever, and Geoffrey~E. Hinton.
\newblock Imagenet classification with deep convolutional neural networks.
\newblock In Fernando Pereira, Christopher~J.C. Burges, Leon Bottou, and
  Kilian~Q. Weinberger, editors, \emph{Advances in neural information
  processing systems (NIPS 2012)}, pages 1097--1105. NIPS, 2012.

\bibitem[Krupinski et~al.(2006)Krupinski, Tillack, Richter, Henderson,
  Bhattacharyya, Scott, Graham, Descour, Davis, and
  Weinstein]{KrupinskiEtAl:2006:EyeMovement}
Elizabeth~A. Krupinski, Allison~A. Tillack, Lynne Richter, Jeffrey~T.
  Henderson, Achyut~K. Bhattacharyya, Katherine~M. Scott, Anna~R. Graham,
  Michael~R. Descour, John~R. Davis, and Ronald~S. Weinstein.
\newblock Eye-movement study and human performance using telepathology virtual
  slides. implications for medical education and differences with experience.
\newblock \emph{Human pathology}, 37\penalty0 (12):\penalty0 1543--1556, 2006.
\newblock \doi{10.1016/j.humpath.2006.08.024}.

\bibitem[Lafon et~al.(2006)Lafon, Keller, and
  Coifman]{LafonEtAl:2006:dataFusionML}
Stephane Lafon, Yosi Keller, and Ronald~R. Coifman.
\newblock Data fusion and multicue data matching by diffusion maps.
\newblock \emph{IEEE Transactions on pattern analysis and machine
  intelligence}, 28\penalty0 (11):\penalty0 1784--1797, 2006.
\newblock \doi{10.1109/TPAMI.2006.223}.

\bibitem[Lake et~al.(2016)Lake, Ullman, Tenenbaum, and
  Gershman]{LakeUlmanTenenbaumGershman2016:MachinesThinkArXiv}
Brenden~M. Lake, Tomer~D. Ullman, Joshua~B. Tenenbaum, and Samuel~J. Gershman.
\newblock Building machines that learn and think like people.
\newblock \emph{arXiv:1604.00289}, 2016.

\bibitem[Lauritzen(1996)]{Lauritzen:1996:GraphicalModels}
Steffen~L. Lauritzen.
\newblock \emph{Graphical models}.
\newblock Clarendon Press, Oxford, 1996.

\bibitem[Le et~al.(2011)Le, Ranzato, Monga, Devin, Chen, Corrado, Dean, and
  Ng]{LeCorradoDeanNg:2011:theCat}
Quoc~V. Le, Marc'Aurelio Ranzato, Rajat Monga, Matthieu Devin, Kai Chen,
  Greg~S. Corrado, Jeff Dean, and Andrew~Y. Ng.
\newblock Building high-level features using large scale unsupervised learning.
\newblock \emph{arXiv:1112.6209}, 2011.

\bibitem[LeCun et~al.(1989)LeCun, Boser, Denker, Henderson, Howard, Hubbard,
  and Jackel]{LeCun:1989:Backprop}
Yann LeCun, Bernhard Boser, John~S. Denker, Donnie Henderson, Richard~E.
  Howard, Wayne Hubbard, and Lawrence~D. Jackel.
\newblock Backpropagation applied to handwritten zip code recognition.
\newblock \emph{Neural computation}, 1\penalty0 (4):\penalty0 541--551, 1989.
\newblock \doi{10.1162/neco.1989.1.4.541}.

\bibitem[LeCun et~al.(2015)LeCun, Bengio, and
  Hinton]{LeCunBengioHinton:2015:DeepLearningNature}
Yann LeCun, Yoshua Bengio, and Geoffrey Hinton.
\newblock Deep learning.
\newblock \emph{Nature}, 521\penalty0 (7553):\penalty0 436--444, 2015.
\newblock \doi{10.1038/nature14539}.

\bibitem[Lee et~al.(2016)Lee, Kim, Lee, and
  Yoon]{LeeYoon:2016:TransferLearningDeepGraphs}
Jaekoo Lee, Hyunjae Kim, Jongsun Lee, and Sungroh Yoon.
\newblock Intrinsic geometric information transfer learning on multiple
  graph-structured datasets.
\newblock \emph{arXiv:1611.04687}, 2016.

\bibitem[Libbrecht and Noble(2015)]{LibbrechtNoble:2015:MLgenetics}
Maxwell~W. Libbrecht and William~Stafford Noble.
\newblock Machine learning applications in genetics and genomics.
\newblock \emph{Nature Reviews Genetics}, 16\penalty0 (6):\penalty0 321--332,
  2015.
\newblock \doi{10.1038/nrg3920}.

\bibitem[Liu et~al.(2017)Liu, Gadepalli, Norouzi, Dahl, Kohlberger, Boyko,
  Venugopalan, Timofeev, Nelson, Corrado, Hipp, Peng, and
  Stumpe]{LiuStumpe:2017:GooglePatho}
Yun Liu, Krishna Gadepalli, Mohammad Norouzi, George~E. Dahl, Timo Kohlberger,
  Aleksey Boyko, Subhashini Venugopalan, Aleksei Timofeev, Philip~Q. Nelson,
  Greg~S. Corrado, Jason~D. Hipp, Lily Peng, and Martin~C. Stumpe.
\newblock Detecting cancer metastases on gigapixel pathology images.
\newblock \emph{arXiv:1703.02442}, 2017.

\bibitem[Long et~al.(2016)Long, Wang, and
  Jordan]{LongWangJordan:2016:DeepTransfer}
Mingsheng Long, Jianmin Wang, and Michael~I. Jordan.
\newblock Deep transfer learning with joint adaptation networks.
\newblock \emph{arXiv:1605.06636}, 2016.

\bibitem[Louw et~al.(2015)Louw, Merat, and Jamson]{LouwEtAl:2015:DriverLoop}
Tyron Louw, Natasha Merat, and Hamish Jamson.
\newblock Engaging with highly automated driving: to be or not to be in the
  loop.
\newblock In \emph{Eighth International Driving Symposium on Human Factors in
  Driver Assessment, Training and Vehicle Design}, pages 190--196, 2015.

\bibitem[Madabhushi and Lee(2016)]{MadabhushiLee:2016:MLDigitalPathology}
Anant Madabhushi and George Lee.
\newblock Image analysis and machine learning in digital pathology: Challenges
  and opportunities.
\newblock \emph{Medical Image Analysis}, 33:\penalty0 170--175, 2016.
\newblock \doi{10.1016/j.media.2016.06.037}.

\bibitem[Malle et~al.(2016)Malle, Kieseberg, Weippl, and
  Holzinger]{MalleEtAl:2016:forgotten}
Bernd Malle, Peter Kieseberg, Edgar Weippl, and Andreas Holzinger.
\newblock The right to be forgotten: Towards machine learning on perturbed
  knowledge bases.
\newblock In \emph{Springer Lecture Notes in Computer Science LNCS 9817}, pages
  251--256. Springer, Heidelberg, Berlin, New York, 2016.
\newblock \doi{10.1007/978-3-319-45507-5_17}.

\bibitem[Malle et~al.(2017)Malle, Giuliani, Kieseberg, and
  Holzinger]{MalleEtAl:2017:FederatedLearning}
Bernd Malle, Nicola Giuliani, Peter Kieseberg, and Andreas Holzinger.
\newblock The more the merrier - federated learning from local sphere
  recommendations.
\newblock In \emph{Machine Learning and Knowledge Extraction, IFIP CD-MAKE,
  Lecture Notes in Computer Science LNCS 10410}, pages 367--374. Springer,
  Cham, 2017.
\newblock \doi{10.1007/978-3-319-66808-6_24}.

\bibitem[McCarthy(2007)]{McCarthy:2007:HumanLevelAI}
John McCarthy.
\newblock From here to human-level ai.
\newblock \emph{Artificial Intelligence}, 171\penalty0 (18):\penalty0
  1174--1182, 2007.
\newblock \doi{10.1016/j.artint.2007.10.009}.

\bibitem[McDermott et~al.(2013)McDermott, Wang, Mitchell, Webb-Robertson,
  Hafen, Ramey, and Rodland]{McDermottEtAl:2013:Biomarker}
Jason~E. McDermott, Jing Wang, Hugh Mitchell, Bobbie-Jo Webb-Robertson, Ryan
  Hafen, John Ramey, and Karin~D. Rodland.
\newblock Challenges in biomarker discovery: combining expert insights with
  statistical analysis of complex omics data.
\newblock \emph{Expert opinion on medical diagnostics}, 7\penalty0
  (1):\penalty0 37--51, 2013.
\newblock \doi{10.1517/17530059.2012.718329}.

\bibitem[Mnih et~al.(2013)Mnih, Kavukcuoglu, Silver, Graves, Antonoglou,
  Wierstra, and Riedmiller]{MnihSilverEtAl:2013:AtariDeepLearningArXiV}
Volodymyr Mnih, Koray Kavukcuoglu, David Silver, Alex Graves, Ioannis
  Antonoglou, Daan Wierstra, and Martin Riedmiller.
\newblock Playing atari with deep reinforcement learning.
\newblock \emph{arXiv:1312.5602}, 2013.

\bibitem[M\"{u}ller et~al.(2015)M\"{u}ller, Reihs, Zatloukal, Jeanquartier,
  Merino-Martinez, van Enckevort, Swertz, and
  Holzinger]{MuellerHolzinger:2015:BiobankIntegration}
Heimo M\"{u}ller, Robert Reihs, Kurt Zatloukal, Fleur Jeanquartier, Roxana
  Merino-Martinez, David van Enckevort, Morris~A. Swertz, and Andreas
  Holzinger.
\newblock State-of-the-art and future challenges in the integration of biobank
  catalogues.
\newblock In Andreas Holzinger, Carsten R\"{o}cker, and Martina Ziefle,
  editors, \emph{Smart Health, Lecture Notes in Computer Science LNCS 8700},
  pages 261--273. Springer, Heidelberg, 2015.
\newblock \doi{10.1007/978-3-319-16226-3_11}.

\bibitem[Ngiam et~al.(2010)Ngiam, Chen, Chia, Koh, Le, and
  Ng]{NigamEtAl:2010:TiledConvolutional}
Jiquan Ngiam, Zhenghao Chen, Daniel Chia, Pang~W. Koh, Quoc~V. Le, and
  Andrew~Y. Ng.
\newblock Tiled convolutional neural networks.
\newblock In \emph{Advances in neural information processing systems (NIPS)},
  pages 1279--1287. NIPS, 2010.

\bibitem[Nie et~al.(2017)Nie, Trullo, Lian, Petitjean, Ruan, Wang, and
  Shen]{NieEtAl:2017:MedicalImageContext}
Dong Nie, Roger Trullo, Jun Lian, Caroline Petitjean, Su~Ruan, Qian Wang, and
  Dinggang Shen.
\newblock Medical image synthesis with context-aware generative adversarial
  networks.
\newblock In \emph{International Conference on Medical Image Computing and
  Computer-Assisted Intervention (MICCAI), Lecture Notes in Computer Science
  LNCS 10435}, pages 417--425. Springer, Cham, 2017.
\newblock \doi{10.1007/978-3-319-66179-7_48}.

\bibitem[O'Sullivan et~al.(2017)O'Sullivan, Holzinger, Zatloukal, Saldiva,
  Sajid, and Wichmann]{Shane:2017:VirtualAutopsyFirst}
Shane O'Sullivan, Andreas Holzinger, Kurt Zatloukal, Paulo Saldiva, Muhammad~I.
  Sajid, and Dominic Wichmann.
\newblock Machine learning enhanced virtual autopsy.
\newblock \emph{Autopsy Case Report}, 7\penalty0 (4):\penalty0 3--7, 2017.
\newblock \doi{10.4322/acr.2017.037}.

\bibitem[Payer et~al.(2016)Payer, \v{S}tern, Bischof, and
  Urschler]{PayerBischof:2016:Landmark}
Christian Payer, Darko \v{S}tern, Horst Bischof, and Martin Urschler.
\newblock Regressing heatmaps for multiple landmark localization using cnns.
\newblock In \emph{International Conference on Medical Image Computing and
  Computer-Assisted Intervention (MICCAI), Springer Lecture Notes in Computer
  Science, LNCS 9901}, pages 230--238. Springer, 2016.
\newblock \doi{10.1007/978-3-319-46723-8_27}.

\bibitem[Pearl(2009)]{Pearl:2009:Causality}
Judea Pearl.
\newblock \emph{Causality: Models, Reasoning, and Inference (Second Edition)}.
\newblock Cambridge University Press, Cambridge, 2009.

\bibitem[Reeder and Felson(2003)]{ReederFelson:2003:Gamuts}
Maurice~Merrick Reeder and Benjamin Felson.
\newblock \emph{Gamuts in radiology: comprehensive lists of roentgen
  differential diagnosis. Fourth Edition.}
\newblock Springer, New York, 2003.

\bibitem[Ribeiro et~al.(2016)Ribeiro, Singh, and
  Guestrin]{RibeiroSinghGuestrin:2016:Trust}
Marco~Tulio Ribeiro, Sameer Singh, and Carlos Guestrin.
\newblock Why should i trust you?: Explaining the predictions of any
  classifier.
\newblock In \emph{22nd ACM SIGKDD International Conference on Knowledge
  Discovery and Data Mining}, pages 1135--1144. ACM, 2016.

\bibitem[Ronneberger et~al.(2015)Ronneberger, Fischer, and
  Brox]{RonnebergerEtAl:2015:Unet}
Olaf Ronneberger, Philipp Fischer, and Thomas Brox.
\newblock U-net: Convolutional networks for biomedical image segmentation.
\newblock In \emph{International Conference on Medical Image Computing and
  Computer-Assisted Intervention (MICCAI), Lecture Notes in Computer Science,
  LNCS 9351}, pages 234--241. Springer, Heidelberg, 2015.
\newblock \doi{10.1007/978-3-319-24574-4_28}.

\bibitem[Roullier et~al.(2011)Roullier, L\'{e}zoray, Ta, and
  Elmoataz]{RoullierEtAl:2011:GraphsWSI}
Vincent Roullier, Olivier L\'{e}zoray, Vinh-Thong Ta, and Abderrahim Elmoataz.
\newblock Multi-resolution graph-based analysis of histopathological whole
  slide images: Application to mitotic cell extraction and visualization.
\newblock \emph{Computerized Medical Imaging and Graphics}, 35\penalty0
  (7):\penalty0 603--615, 2011.
\newblock \doi{10.1016/j.compmedimag.2011.02.005}.

\bibitem[Rozantsev et~al.(2015)Rozantsev, Lepetit, and
  Fua]{RozantsevEtAl:2015:RenderingImages}
Artem Rozantsev, Vincent Lepetit, and Pascal Fua.
\newblock On rendering synthetic images for training an object detector.
\newblock \emph{Computer Vision and Image Understanding}, 137:\penalty0 24--37,
  2015.
\newblock \doi{10.1016/j.cviu.2014.12.006}.

\bibitem[Russell and Norvig(1995)]{RusselNorvig:1995:AIbook}
Stuart~J. Russell and Peter Norvig.
\newblock \emph{Artificial Intelligence: A modern approach}.
\newblock Prentice Hall, Englewood Cliffs, 1995.

\bibitem[Saul and Roweis(2003)]{SaulRoweis:2003:Manifold}
Lawrence~K. Saul and Sam~T. Roweis.
\newblock Think globally, fit locally: unsupervised learning of low dimensional
  manifolds.
\newblock \emph{Journal of machine learning research (JMLR)}, 4:\penalty0
  119--155, 2003.

\bibitem[Schmidhuber(2015)]{Schmidhuber:2015:DLOverview}
J\"{u}rgen Schmidhuber.
\newblock Deep learning in neural networks: An overview.
\newblock \emph{Neural Networks}, 61\penalty0 (1):\penalty0 85--117, 2015.
\newblock \doi{10.1016/j.neunet.2014.09.003}.

\bibitem[Silver et~al.(2016)Silver, Huang, Maddison, Guez, Sifre, van~den
  Driessche, Schrittwieser, Antonoglou, Panneershelvam, Lanctot, Dieleman,
  Grewe, Nham, Kalchbrenner, Sutskever, Lillicrap, Leach, Kavukcuoglu, Graepel,
  and Hassabis]{HassabisEtAl:2016:GoNature}
David Silver, Aja Huang, Chris~J. Maddison, Arthur Guez, Laurent Sifre, George
  van~den Driessche, Julian Schrittwieser, Ioannis Antonoglou, Veda
  Panneershelvam, Marc Lanctot, Sander Dieleman, Dominik Grewe, John Nham, Nal
  Kalchbrenner, Ilya Sutskever, Timothy Lillicrap, Madeleine Leach, Koray
  Kavukcuoglu, Thore Graepel, and Demis Hassabis.
\newblock Mastering the game of go with deep neural networks and tree search.
\newblock \emph{Nature}, 529\penalty0 (7587):\penalty0 484--489, 2016.
\newblock \doi{10.1038/nature16961}.

\bibitem[Silver et~al.(2017)Silver, Schrittwieser, Simonyan, Antonoglou, Huang,
  Guez, Hubert, Baker, Lai, Bolton, Chen, Lillicrap, Hui, Sifre, Driessche,
  Graepel, and Hassabis]{SilverHassabisEtAl:2017:GoWithoutHuman}
David Silver, Julian Schrittwieser, Karen Simonyan, Ioannis Antonoglou, Aja
  Huang, Arthur Guez, Thomas Hubert, Lucas Baker, Matthew Lai, Adrian Bolton,
  Yutian Chen, Timothy Lillicrap, Fan Hui, Laurent Sifre, George van~den
  Driessche, Thore Graepel, and Demis Hassabis.
\newblock Mastering the game of go without human knowledge.
\newblock \emph{Nature}, 550\penalty0 (7676):\penalty0 354--359, 2017.
\newblock \doi{doi:10.1038/nature24270}.

\bibitem[Singh et~al.(2016)Singh, Ribeiro, and
  Guestrin]{SinghRibeiroGuestrin:2016:BlackBoxArxiv}
Sameer Singh, Marco~Tulio Ribeiro, and Carlos Guestrin.
\newblock Programs as black-box explanations.
\newblock \emph{arXiv:1611.07579}, 2016.

\bibitem[Strogatz(2001)]{Strogatz2001complexNetworks}
Steven~H. Strogatz.
\newblock Exploring complex networks.
\newblock \emph{Nature}, 410\penalty0 (6825):\penalty0 268--276, 2001.
\newblock \doi{10.1038/35065725}.

\bibitem[Swan et~al.(2013)Swan, Mobasheri, Allaway, Liddell, and
  Bacardit]{SwanEtAl:2013:MLProteomics}
Anna~Louise Swan, Ali Mobasheri, David Allaway, Susan Liddell, and Jaume
  Bacardit.
\newblock Application of machine learning to proteomics data: Classification
  and biomarker identification in postgenomics biology.
\newblock \emph{Omics-a Journal of Integrative Biology}, 17\penalty0
  (12):\penalty0 595--610, 2013.
\newblock \doi{10.1089/omi.2013.0017}.

\bibitem[Tenenbaum et~al.(2000)Tenenbaum, de~Silva, and
  Langford]{TenenbaumSilvaLangford:2000:Isomap}
Joshua~B. Tenenbaum, Vin de~Silva, and John~C. Langford.
\newblock A global geometric framework for nonlinear dimensionality reduction.
\newblock \emph{Science}, 290\penalty0 (5500):\penalty0 2319--2323, 2000.
\newblock \doi{10.1126/science.290.5500.2319}.

\bibitem[Tran et~al.(2017)Tran, Hoffman, Saurous, Brevdo, Murphy, and
  Blei]{TranBlei:2017:DeepProbProg}
Dustin Tran, Matthew~D. Hoffman, Rif~A. Saurous, Eugene Brevdo, Kevin Murphy,
  and David~M. Blei.
\newblock Deep probabilistic programming.
\newblock \emph{arXiv:1701.03757}, 2017.

\bibitem[Van~Dyk and Meng(2001)]{VanDykMeng:2001:DataAugmentation}
David~A. Van~Dyk and Xiao-Li Meng.
\newblock The art of data augmentation.
\newblock \emph{Journal of Computational and Graphical Statistics}, 10\penalty0
  (1):\penalty0 1--50, 2001.
\newblock \doi{10.1198/10618600152418584}.

\bibitem[van Ommen et~al.(2015)van Ommen, T\"{o}rnwall, Br\'{e}chot, Dagher,
  Galli, Hveem, Landegren, Luchinat, Metspalu, Nilsson, Solesvik, Perola,
  Litton, and Zatloukal]{OmmenZatloukal:2015:BBMRI}
Gert-Jan~B. van Ommen, Outi T\"{o}rnwall, Christian Br\'{e}chot, Georges
  Dagher, Joakim Galli, Kristian Hveem, Ulf Landegren, Claudio Luchinat, Andres
  Metspalu, Cecilia Nilsson, Ove~V Solesvik, Markus Perola, Jan-Eric Litton,
  and Kurt Zatloukal.
\newblock {BBMRI-ERIC} as a resource for pharmaceutical and life science
  industries: the development of biobank-based expert centres.
\newblock \emph{European Journal of Human Genetics}, 23\penalty0 (7):\penalty0
  893--900, 2015.
\newblock \doi{10.1038/ejhg.2014.235}.

\bibitem[Villaronga et~al.(2017)Villaronga, Kieseberg, and
  Li]{VillarongaEtAl:2017:HumansForget}
Eduard~Fosch Villaronga, Peter Kieseberg, and Tiffany Li.
\newblock Humans forget, machines remember: Artificial intelligence and the
  right to be forgotten.
\newblock \emph{Computer Law \& Security Review}, 2017.
\newblock \doi{10.1016/j.clsr.2017.08.007}.

\bibitem[Virchow(1871)]{Virchow:1871:Patho}
Rudolf Virchow.
\newblock \emph{Vorlesungen ü\"{u}ber Pathologie I: Die Cellular-Pathologie in
  ihrer Begrü\"{u}ndung auf physiologische und pathologische Gewebelehre. 4.
  Auflage}.
\newblock August Hirschwald, Berlin, 1871.

\bibitem[Weber et~al.(2017)Weber, Racani\`{e}re, Reichert, Buesing, Guez,
  Rezende, Badia, Vinyals, Heess, and Li]{WeberEtAl:2017:AugmentedAgentsDeep}
Th\'{e}ophane Weber, S\'{e}bastien Racani\`{e}re, David~P. Reichert, Lars
  Buesing, Arthur Guez, Danilo~Jimenez Rezende, Adria~Puigdomènech Badia,
  Oriol Vinyals, Nicolas Heess, and Yujia Li.
\newblock Imagination-augmented agents for deep reinforcement learning.
\newblock \emph{arXiv:1707.06203}, 2017.

\bibitem[Weinberger and Saul(2006)]{WeinbergerSaul:2006:Manifolds}
Kilian~Q. Weinberger and Lawrence~K. Saul.
\newblock Unsupervised learning of image manifolds by semidefinite programming.
\newblock \emph{International journal of computer vision (IJCV)}, 70\penalty0
  (1):\penalty0 77--90, 2006.
\newblock \doi{10.1007/s11263-005-4939-z}.

\bibitem[Weinstein et~al.(1987)Weinstein, Bloom, and
  Rozek]{Weinstein:1987:telepathology}
Ronald~S. Weinstein, Ken~J. Bloom, and Laura~S. Rozek.
\newblock Telepathology and the networking of pathology diagnostic services.
\newblock \emph{Archives of pathology \& laboratory medicine}, 111\penalty0
  (7):\penalty0 646--652, 1987.

\bibitem[Weinstein et~al.(2009)Weinstein, Graham, Richter, Barker, Krupinski,
  Lopez, Erps, Bhattacharyya, Yagi, and
  Gilbertson]{WeinsteinEtAl:2009:VirtualMicroscopy}
Ronald~S. Weinstein, Anna~R. Graham, Lynne~C. Richter, Gail~P. Barker,
  Elizabeth~A. Krupinski, Ana~Maria Lopez, Kristine~A. Erps, Achyut~K.
  Bhattacharyya, Yukako Yagi, and John~R. Gilbertson.
\newblock Overview of telepathology, virtual microscopy, and whole slide
  imaging: prospects for the future.
\newblock \emph{Human pathology}, 40\penalty0 (8):\penalty0 1057--1069, 2009.
\newblock \doi{10.1016/j.humpath.2009.04.006}.

\bibitem[Zeiler and Fergus(2013)]{ZeilerFergus:2013:VisDeepArXiV}
Matthew~D. Zeiler and Rob Fergus.
\newblock Visualizing and understanding convolutional networks.
\newblock \emph{arXiv:1311.2901}, 2013.

\bibitem[Zeiler et~al.(2011)Zeiler, Taylor, and
  Fergus]{ZeilerTaylorFergus:2011:deconvnet}
Matthew~D. Zeiler, Graham~W. Taylor, and Rob Fergus.
\newblock Adaptive deconvolutional networks for mid and high level feature
  learning.
\newblock In \emph{IEEE International Conference on Computer Vision (ICCV)},
  pages 2018--2025. IEEE, 2011.
\newblock \doi{10.1109/ICCV.2011.6126474}.

\bibitem[Zhang et~al.(2009)Zhang, Song, Gretton, and
  Smola]{ZhangEtAl:2009:nonIID}
Xinhua Zhang, Le~Song, Arthur Gretton, and Alex~J. Smola.
\newblock Kernel measures of independence for non-iid data.
\newblock In \emph{Advances in neural information processing systems (NIPS)},
  pages 1937--1944, 2009.
\newblock URL
  \url{http://papers.nips.cc/paper/3440-kernel-measures-of-independence-for-non-iid-data}.

\bibitem[Zomorodian(2010)]{Zomorodian:2009:COMPTOPinHandbook}
Afra Zomorodian.
\newblock Computational topology.
\newblock In M.~Atallah and M.~Blanton, editors, \emph{Algorithms and Theory of
  Computation Handbook, Second Edition, Volume 2: Special Topics and
  Techniques}, pages 1--31. Chapman and Hall/CRC, Boca Raton (FL), 2010.
\newblock \doi{doi:10.1201/9781584888215-c3}.

\bibitem[Zomorodian and
  Carlsson(2005)]{ZomorodianCarlsson:2005:PersistentHomology}
Afra Zomorodian and Gunnar Carlsson.
\newblock Computing persistent homology.
\newblock \emph{Discrete \& Computational Geometry}, 33\penalty0 (2):\penalty0
  249--274, 2005.
\newblock \doi{10.1007/s00454-004-1146-y}.

\end{thebibliography}

\end{document}